\documentclass[journal]{IEEEtran}

\usepackage{graphicx,subfigure,epsfig,amsmath,amssymb,multirow,times}
\usepackage{algorithmic,balance,cite}
\usepackage{algorithm}
\usepackage[update,prepend]{epstopdf}
\usepackage{color}

\newtheorem{thm}{Theorem}[section]
\newtheorem{defn}{Definition}[section]

\ifCLASSINFOpdf
\else
\fi
\begin{document}
%
\title{Scene Image is Non-Mutually Exclusive - \\ A Fuzzy Qualitative Scene Understanding}
%
%
%
\author{Chern~Hong~Lim,~\IEEEmembership{Student Member,~IEEE,}
        Anhar~Risnumawan,~\IEEEmembership{Student Member,~IEEE,}
       and~Chee~Seng~Chan,~\IEEEmembership{Member,~IEEE}
\thanks{Manuscript received May 09, 2013, revised September 19, 2013. Accepted for publication November 28, 2013. This work is supported by the University of Malaya HIR under grant No UM.C/625/1/HIR/037, J0000073579}
\thanks{Authors are with the Centre of Image and Signal Processing, Faculty of Computer Science and Information Technology, University of Malaya, 50603 Kuala Lumpur, MALAYSIA. Corresponding e-mail: (cs.chan@um.edu.my).}}
        

%
%

\markboth{{\tiny This article has been accepted for publication in a future issue of IEEE Trans. Fuzzy Sys., but has not been fully edited. Content may change prior to final publication.}}%
{}
%

\IEEEpubid{{\tiny Copyright (c) 2014 IEEE. Personal use is permitted. For any other purposes, permission must be obtained from the IEEE by emailing pubs-permissions@ieee.org.}}


\maketitle

\begin{abstract}

Ambiguity or uncertainty is a pervasive element of many real world decision making processes. Variation in decisions is a norm in this situation when the same problem is posed to different subjects. Psychological and metaphysical research had proven that decision making by human is subjective. It is influenced by many factors such as experience, age, background, etc. Scene understanding is one of the computer vision problems that fall into this category. Conventional methods relax this problem by assuming scene images are mutually exclusive; and therefore, focus on developing different approaches to perform the binary classification tasks. In this paper, we show that scene images are non-mutually exclusive, and propose the Fuzzy Qualitative Rank Classifier (FQRC) to tackle the aforementioned problems. The proposed FQRC provides a ranking interpretation instead of binary decision. Evaluations in term of qualitative and quantitative using large numbers and challenging public scene datasets have shown the effectiveness of our
proposed method in modeling the non-mutually exclusive scene images. 
\end{abstract}



%
\IEEEpeerreviewmaketitle

\section{Introduction}

\begin{figure}[tb]
\centering
\subfigure[Coast]{\includegraphics[scale=0.35]{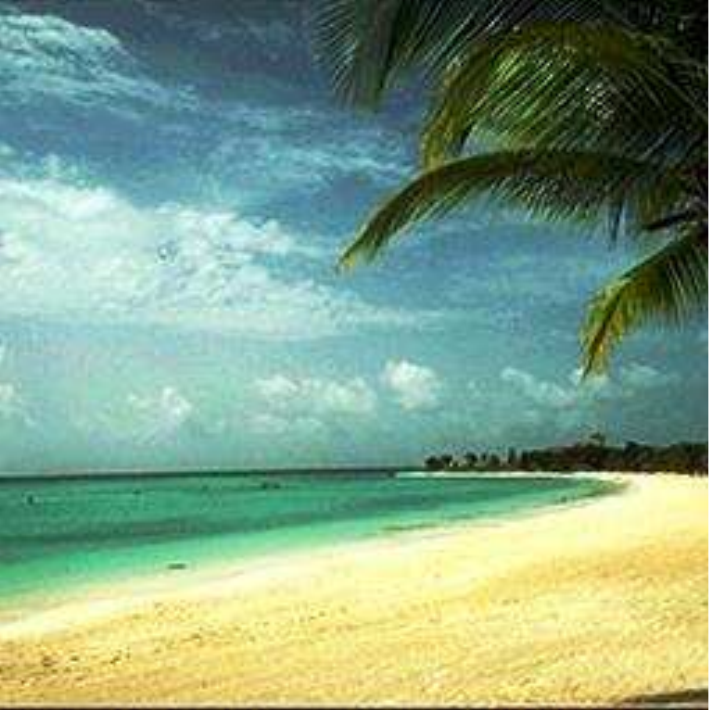}
\label{fig:Coast_intro}}
\subfigure[?]{\includegraphics[scale=0.35]{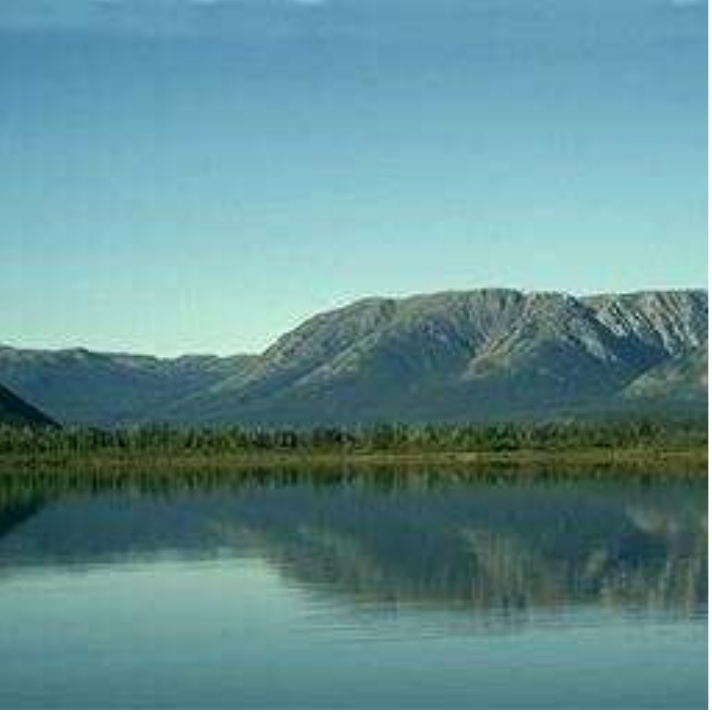}
\label{fig:Coast_Mountain_intro}}
\subfigure[Mountain]{\includegraphics[scale=0.35]{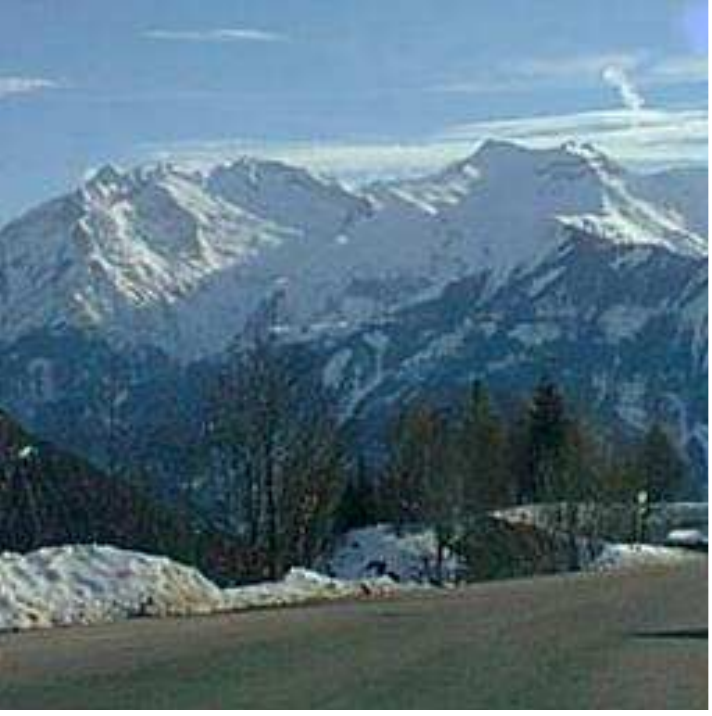}
\label{fig:Mountain_intro}}
\caption{Example of ambiguous scene between Coast and Mountain.}
\label{fig:AmPic1}
\end{figure}

One of the biggest challenges in real world decision making process is to cope with uncertainty, complexity, volatility and ambiguity. How do we deal with this growing confusion in our world? In scene understanding, an important and yet difficult image understanding problem due to their variability, ambiguity, wide range of illumination and scale conditions falls into this category. The conventional goal of the works is to assign an unknown scene image to one of the several possible classes. For example, Fig. \ref{fig:Coast_intro} is a Coast class scene while Fig. \ref{fig:Mountain_intro} is a Mountain class scene. 

Intentionally, most state-of-the-art approaches in scene understanding domain \cite{Oliva_Torralba_2001,Fei-Fei_Perona_2005,Bosch_Zisserman_Mu_2006,Vogel_Schiele_2007} are exemplar-based and assume that scene images are mutually exclusive, $P(A \, \cap \, B)=0$.  This simplifies the complex problem of scene understanding (uncertainty, complexity, volatility, and ambiguity) to a simple binary classification task. Such approaches learn patterns from a training set and subsequently, search for the images similar to it. As a result of this, classification errors often occur when the scene classes overlap in the selected feature space. For example, it is unclear that in Fig. \ref{fig:Coast_Mountain_intro} is a Coast class scene or a Mountain class scene. 

Inspired by the fuzzy set theory proposed by Lotfi Zadeh \cite{Zadeh1965338}, we argue that scene images are {\bf non-mutually exclusive} where different people are likely to respond inconsistently. Here, we define inconsistent as the scenario where there is no definite answer (in computational term, a binary or linear answer) to a problem. This notion became popular among researchers and technologists due to wide spectrum of applications \cite{du2009theory,qiu2009fuzzy}. In scene understanding, however, only a few numbers of the research works are aware of and had tackled this problem. The notable ones are \cite{chernhong,Boutell_Luo_Shen_Brown_2004, Zhang_2007}, where a multi-label scene classification framework is proposed. However, these approaches are not practical due to: firstly, the work requires human intervention to manually annotate the multi-label training data. This is a tedious job that leads to a large number of classes with the sparse number of sample \cite{tsoumakas2007multi}. Secondly, the annotated image's classes are potentially bias as different people tend to respond inconsistently \cite{Forguson} and finally, it does not able to handle multi-dimension data.

In this paper, our aim is to study a novel approach to remedy the aforementioned problems. We propose the Fuzzy Qualitative Rank-Classifier (FQRC) to relax the assumption that scene images are mutually exclusive. Therefore, a scene can be somewhat arbitrary and possibly sub-optimal. We compare the results from FQRC with an online survey to show that there is an influence of human factors (background, experience, age, etc.) in decision making and hence conclude that assuming scene images are mutually exclusive is impractical. Qualitative and quantitative comparisons to the state-of-the-art solutions have shown the strength and effectiveness of our proposed method. 

\IEEEpubidadjcol
In summary, our main contribution is to show that scene images are
non-mutually exclusive. This is supported by an online survey that
participated by more than 150 candidates from different ethnics and age using the OSR dataset \cite{Oliva_Torralba_2001}. The reason is to raise the awareness of computer vision community regarding this very important, but largely neglected issue. With this in mind, we propose the FQRC to model scene images in a non-mutually exclusive manner, where we develop an inference that outputs ranking result. In advance, FQRC provides a resolution toward conventional solutions which either perform binary classification
\cite{Oliva_Torralba_2001,Fei-Fei_Perona_2005,Bosch_Zisserman_Mu_2006,
  Vogel_Schiele_2007} or require human intervention \cite{Boutell_Luo_Shen_Brown_2004,Zhang_2007}.

The rest of the paper is organized as follows. Section \ref{RW} covers the related works in scene understanding. Section \ref{FQC} presents our proposed framework which consists of two stages, the learning and inference stages. The intuition and stability analysis of our proposed approach are discussed in Section \ref{disOfRQRC}. Section \ref{IR} demonstrates the ranking interpretation. Section \ref{ExpSet} shows the experiment results, and finally, we conclude in Section \ref{Con}.

\section{Related Work}
\label{RW}

Scene understanding has been one of the mainstream tasks in computer vision. It differs from the conventional object detection or classification tasks, to the extent that a scene is composed of several entities that are often organized in an unpredictable layout \cite{Quelhas}. Surprisingly from our findings, there is very minimal or almost none that had tackled this problem using the fuzzy approach. The early efforts in this area were dominated by computer vision researchers who focus on using machine learning techniques. These prior works denoted the scene understanding problem were to assign one of the several possible classes to a scene image of unknown class.

Oliva and Torralba \cite{Oliva_Torralba_2001} proposed a set of perceptual dimensions (naturalness, openness, roughness, expansion, ruggedness) that represents the dominant spatial structure of a scene - the spatial envelope as scene representation. Then, a support vector machine (SVM) classifier with Gaussian kernel is employed to classify the scene classes. Fei-Fei and Perona \cite{Fei-Fei_Perona_2005} proposed the Bayesian hierarchical model extended from latent dirichlet allocation (LDA) to learn natural scene categories. In their learning model, they represent the image of a scene by a collection of local regions, denoted as codewords obtained by unsupervised learning, finally they choose the best model as their classification result. Bosch et al. \cite{Bosch_Zisserman_Mu_2006} inspired from the previous work and proposed probabilistic latent semantic analysis (pLSA) incorporate with KNN for scene classification. Vogel and Schiele \cite{Vogel_Schiele_2007} used the occurring frequency of different concepts (water, rock, etc.) in an image as the intermediate features for scene image classification. The two-stage system makes use of an intermediary semantic level of block classification (concept level) to do retrieval based on the occurrence of such concepts in an image. 

However, in scene classification task, it is very likely that a scene image can belongs to multiple classes. As a result of this, all the aforementioned solutions that assumed scene classes are mutually exclusive are not practical and often lead to classification errors. We believe that scene images are somewhat arbitrary and possibly sub-optimal as depicted in Fig. \ref{fig:AmPic1}. To the best of our knowledge, there are numerous multi-label classification research \cite{tsoumakas2007multi,tsoumakas2010mining}; however, only a few were focused in the domain of scene understanding. Boutell et al. \cite{Boutell_Luo_Shen_Brown_2004} proposed an approach using SVM with cross-training to build the classifier for every base class. Then maximum a posteriori (MAP) principle is applied with the aid of prior probability calculation and gamma fit operation toward the single and multi-label training data. This is to obtain the desired threshold to determine whether a testing sample is fall into single label event or multiple label events.

Inspired by \cite{Boutell_Luo_Shen_Brown_2004}, Zhang and Zhou \cite{Zhang_2007} introduced multi-label lazy learning K-nearest neighbor (ML-KNN) as their classification algorithm. This is to resolve the inefficiency of using multiple independent binary classifier for each class by using SVM. Statistical information from the training set and MAP principle is utilized to determine the best label for the test instance. Unfortunately, both these methods required manual human annotation of multi-label class training data to compute the prior probability based on frequency counting of training set. This is an impractical solution since a human decision is bias and inconsistent. It also leads to large number of classes with sparse sample \cite{tsoumakas2007multi}. Besides that, human reasoning does not annotate an image as multi-class. For instance, referring to Fig. \ref{fig:Coast_Mountain_intro}, it is very rare for one to say that ``this is a Coast + Mountain class scene image''. In general, one would rather comment ``this is a Coast'' or ``this is a Mountain'' scene.

In what constitutes the closer work to ours in the fuzzy domain, Lim and Chan \cite{chernhong} proposed a fuzzy qualitative framework and Cho and Chang \cite{chang1} employed a simple fuzzy logic with two monocular images to understand the scene images. However, their work suffered from 1) finding the appropriate resolution to build their 4-tuple membership function. Currently, the model parameters are chosen manually based on prior information and in a trial-and-error manner. This is a very tedious and time consuming approach; 2) only able to accommodate two feature vectors as input data; 3) the ranking is undefined and finally 4) tested on a very limited and easy dataset (a dataset that contains only 2 scene images).

In this paper, we extend the work of \cite{chernhong} by learning the 4-tuple membership function from the training data. In order to achieve this, we used the histogram representation. It relaxes the difficulty of obtaining multi-label training data as to \cite{Boutell_Luo_Shen_Brown_2004,Zhang_2007} where the training steps require human intervention in manually annotate the multi-label training data. This is a daunting task as human decisions are subjective and huge amount of participants are needed. Besides that, a ranking method to describe the relationship of image to each scene class is introduced.  In scene understanding, in particular where we model the scene images as non-mutually exclusive, the idea of inference engine with ranking interpretation is somehow new and unexplored.

\section{Fuzzy Qualitative Rank Classifier}
\label{FQC}

\begin{figure}[ht]
\centering
\includegraphics[scale=0.3]{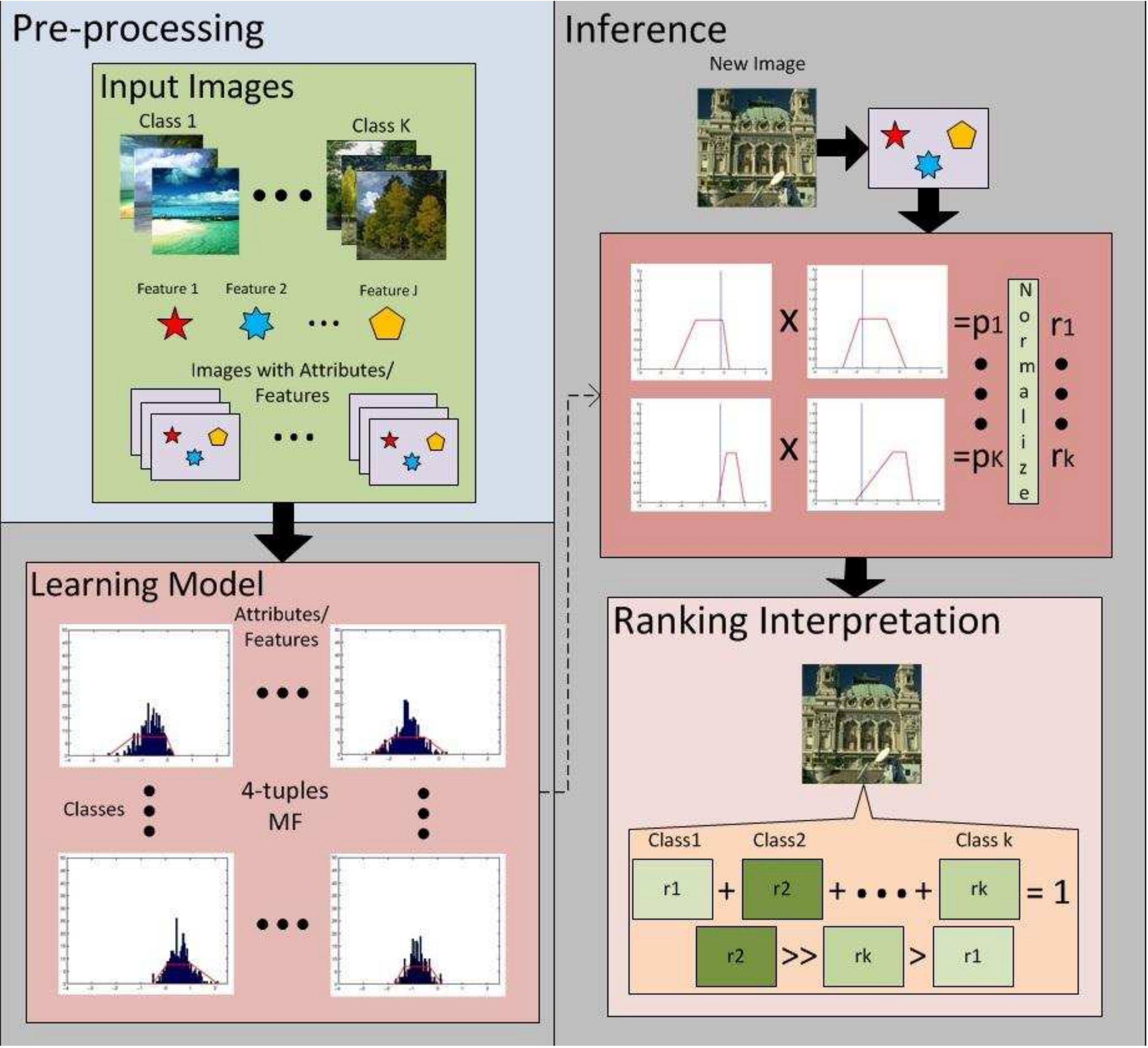}
\caption{Overall framework of the proposed Fuzzy Qualitative Rank Classifier which consists of Pre-processing, Learning Model, Inference, and Ranking Interpretation. }
\label{fig:FQ_BlockDiagram}
\end{figure}

\subsection{Basic Notation}

The general framework of the proposed FQRC consists of four stages: 1) Pre-processing; 2) Learning model; 3) Inference and 4) Ranking interpretation as illustrated in Fig. \ref{fig:FQ_BlockDiagram}. Let $\mathbf{I}=\{\mathbf{I}_1,\mathbf{I}_2,\ldots,\mathbf{I}_N\}$ denotes the $N$ scene images. During the pre-processing stage, any existing feature representation such as texture component, color spectrum and interest point can be employed as an input to our learning model. In this paper, we have employed the attributes \cite{Parikh_Grauman_2011} as our image features. Let $T$ denotes a feature extraction function, $T:\mathbf{I} \to \mathbf{x}_k$, where $\mathbf{x}_k$ is a set of feature values belong to the $k$-th class, $k\in \{1,2,\ldots,K\}$, of input space $\mathcal{X}$, $K$ is the number of classes label. Input data, $\mathbf{x}_k \in \mathcal{X}$, is defined as $\mathbf{x}_k=\{x_1,x_2,\ldots,x_J\}_{k}$, where $x_j$ is a feature value, $j\in \{1,2,\ldots,J\}$, and $J$ is the number of features. We denote sample $(\textbf{x},y)$ as $z\in \mathcal{Z}$ of sample space $\mathcal{Z}$. 

\subsection{Motivation}
In general, the task of a classifier (we denote it as a function $f$) is to find a way, which, based on the observations, assigns a sample to a specified class label, $y\in (\mathcal{Y} \subseteq \{1,2,\ldots,K\})$, where $\mathcal{Y}$ is the output space. The task is to estimate a function $(f \in \mathcal{F}) :\mathbf{x} \to y$, where $\mathcal{F}$ is the function space. A function $f$ is i.i.d., generated using the input-output pairs according to an unknown distribution $P(\mathbf{x},y)$ so that $f$ can classify unseen samples $(\mathbf{x},y)$,

\begin{equation}
(\mathbf{x}_1,y_1),\ldots,(\mathbf{x}_N,y_N)\in (\mathcal{X} \times \mathcal{Y})^N
\end{equation}

The best function $f$, which one can obtain is the one that minimizes the bound of error represented by a risk function (\ref{eq:Rf}). However, one must note that, we could not directly compute the risk $R(f)$ since the probability of
$P(\mathbf{x}, y)$ is unknown.

\begin{equation}
R(f)=\int loss(f(\mathbf{x}),y) P(\mathbf{x}, y)
\label{eq:Rf}
\end{equation}

In scene understanding, (\ref{eq:Rf}) is much difficult to achieve since scene images are non-mutually exclusive due to the inconsistent of human decision, where different people tend to provide different answers. Theoretically, the importance of the non-mutually exclusive data can be derived from the inequality Chernoff bound \cite{chernoff1952measure}:

\begin{equation}
P\left \{ \left | \frac{1}{N}\sum_{i=1}^{N}\mathbf{x}_i-E[\mathbf{x}] \right | \geq \epsilon \right \} \leq 2\exp(-2N\epsilon^2)
\label{eq:Chernoff_1}
\end{equation}

\noindent This theorem states that the probability of sample mean differ by more than $\epsilon$ from the expected mean is bounded by the exponential that depends on the number of samples $N$. Note that if we have more data, the probability of deviation error will converge to zero. However, this is not true because of uniform convergence of function space $\mathcal{F}$ \cite{von2008statistical}. Using the risk function (\ref{eq:Rf}) we can represent the inequality (\ref{eq:Chernoff_1}) as follows,

\begin{equation}
P\left \{ \left | R_{emp}(f)-R(f) \right | \geq \epsilon \right \} \leq 2\exp(-2N\epsilon^2)
\label{eq:Chernoff_2}
\end{equation}

\noindent where $R_{emp}(f)$ and $R(f)$ are the empirical and actual risk, respectively. Inequality (\ref{eq:Chernoff_2}) shows that for a certain function $f$ it is highly probable that the empirical error provides good estimates of the actual risk. Luxburg and Scholkopf \cite{von2008statistical} stated that the empirical risk $R_{emp}(f)$ can be inaccurate when $N \rightarrow \infty$ since Chernoff bound only holds for a fixed function $f$ which does not depend on the training data. But in contrary, $f$ does depend on training data. Therefore, they came up with the uniform convergence and obtained the following inequality:

\begin{equation}
P\left \{ \underset {f\in \mathcal{F}}{\sup}\left | R_{emp}(f)-R(f) \right | \geq \epsilon \right \} \leq 2\exp(-2N\epsilon^2)
\label{eq:Chernoff_3}
\end{equation}

Suppose we have finitely $g$ functions,
$\mathcal{F}=\{f_1,f_2,\ldots,f_g\}$ and $\mathcal{C}^i=\left |
  R_{emp}(f_i)-R(f_i) \right | \geq \epsilon$, then using the union bound we can represent (\ref{eq:Chernoff_3}) as:

\begin{equation}
\begin{aligned}
& P\left \{\underset {f\in \mathcal{F}}{\sup}\left | R_{emp}(f)-R(f) \right | \geq \epsilon \right \} \\ 
&= P(\mathcal{C}^1 \vee \mathcal{C}^2 \vee \cdots \vee \mathcal{C}^{g} ) \\
&= \sum_{i=1}^{g}P(\mathcal{C}^i) - 
\left \{ \mathcal{D}^2 + \mathcal{D}^3 + \cdots + \mathcal{D}^{g} \right \} \\
&\leq \underbrace{ 2g\exp(-2N\epsilon^2) }_\text{1st term} - \underbrace{ \text{bound}( \mathcal{D}^2 + \mathcal{D}^3 + \cdots + \mathcal{D}^{g}) }_\text{2nd term}
\label{eq:Chernoff_4}
\end{aligned}
\end{equation}

\noindent where $\mathcal{D}^{i}$ is the sum of the probabilities of every combination of $i$ event, e.g, $\mathcal{D}^{g}=P(\mathcal{C}^1 \wedge \mathcal{C}^2 \wedge \cdots \wedge \mathcal{C}^g)$. This leads to a bound which states that the probability that empirical risk is close to the actual risk is upper bounded by two terms. The first term is the error bound because of the mutually exclusive data and the second term is due to the non-mutually exclusive data. Most of the conventional classification methods, however, only utilize the mutually exclusive part. In contrast, our proposed method - the FQRC models both the mutually and non-mutually exclusive parts. 

\subsection{Learning the FQRC}
\label{Mod}

In our learning model, we learn the non-mutually exclusive scene data with parametric approximation of the membership function where the membership distribution of a normal convex fuzzy number is approximated by the 4-tuple. This fuzzy representation of qualitative values is more general than ordinary (crisp) interval representations, since it can represents not only the information stated by a well-determined real interval but also the knowledge embedded in the soft boundaries of the interval \cite{Liu_Coghill_Barnes_2009}. Thus, the fuzzy representation removes, or largely weakens (if not completely resolving), the boundary interpretation problem, achieved through the description of a gradual rather than an abrupt change in the degree of membership of which a physical quantity is mapped onto a particular qualitative value. It is, therefore, closer to the common sense intuition of the non-mutually exclusive problem. 

\begin{figure}[tbp]
\centering
\includegraphics[scale=0.3]{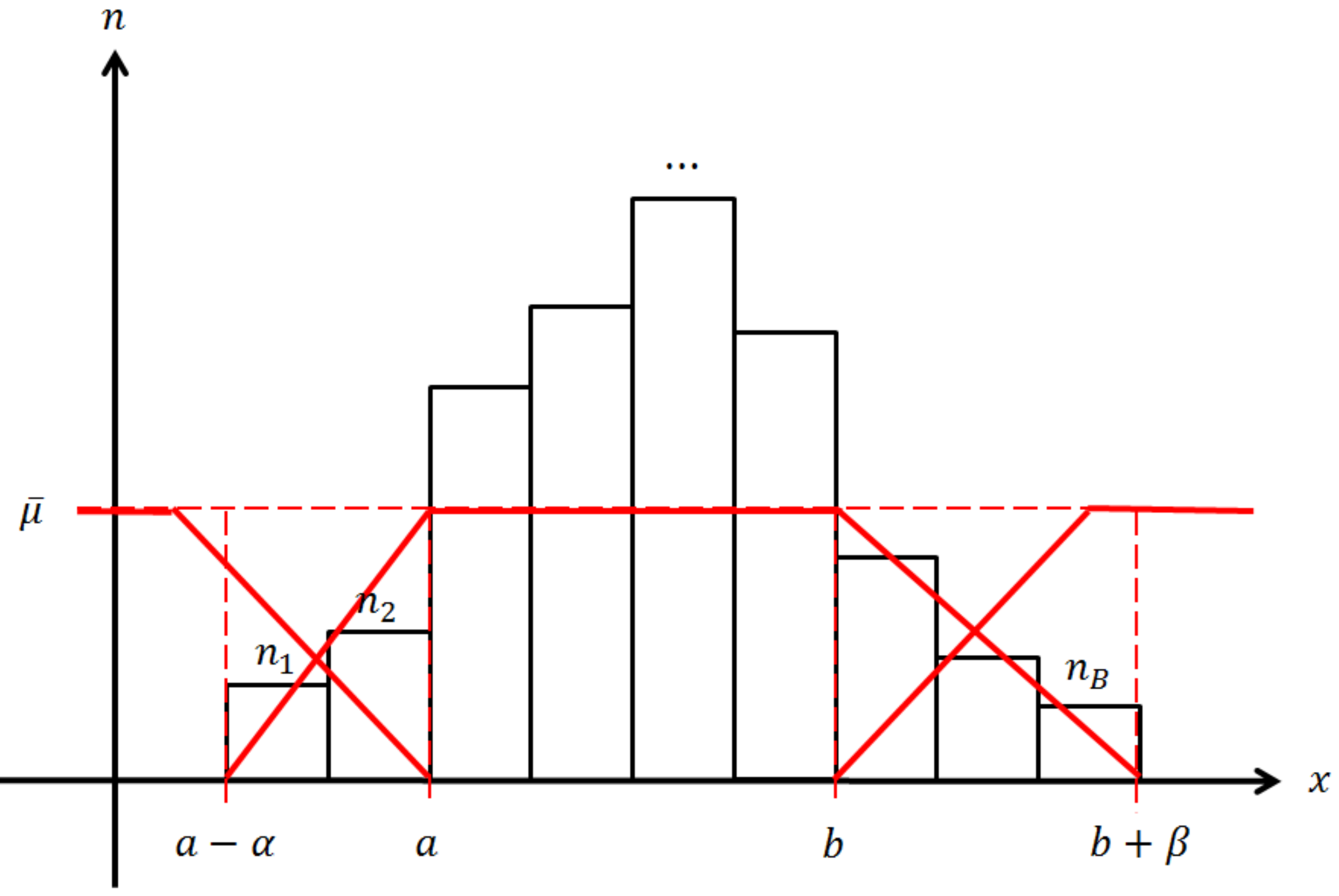}
\caption[membership function]{Parametric representation of a histogram, $x$ is the feature value, $n$ denotes the occurrence of training data from its respective bin $n_1, n_2,\ldots,n_B$. $a$ and $b$ represent the lower and upper bound of $\bar \mu$, while $a-\alpha$ and $b+\beta$ represent the minimum and maximum of $x$ value. The dominant region (mutually exclusive) is the area of $[a,b]$. The intersection area (non-mutually exclusive) is the areas of $[a-\alpha,a]$ and $[b,b+\beta]$.}
\label{fig:mf}
\end{figure}

According to
\cite{Shen_Leitch_1993,Brown_Coghill_2008,Liu_2008,Chan_Liu_2009,ChanLB07,Liu_Coghill_Barnes_2009,ChanLBK08}, such representation of the 4-tuple fuzzy number is a better qualitative representation as the representation has high resolution and good compositionality. The dominant region of 4-tuple indicates the mutually exclusive part, while the intersection between 4-tuple indicates the non-mutually exclusive, as shown in Fig. \ref{fig:mf}. The 4-tuple fuzzy number is represented as
$\mathbf{m}=\{a,b,\alpha,\beta\}$ with the condition $a < b$ and $ab >
0$. There will be $J \times K$ matrix containing 4-tuple for each
feature number and class, as in (\ref{eq:CMF}). Those 4-tuples are represented in the form as $\mathbf{m}_{jk} = \{a,b,\alpha,\beta\}_{jk}$.

\begin{equation}
\mathbf{M} =\begin{bmatrix}
 \mathbf{m}_{11} & \mathbf{m}_{12} & \cdots & \mathbf{m}_{1K}\\ 
 \mathbf{m}_{21} & \mathbf{m}_{22} & \cdots & \mathbf{m}_{2K}\\ 
 \vdots & \vdots & \ddots & \vdots \\ 
 \mathbf{m}_{J1} & \mathbf{m}_{J2} & \cdots & \mathbf{m}_{JK}
\end{bmatrix}
\label{eq:CMF}
\end{equation}

The representation in (\ref{eq:CMF}) is to conserve the appropriate membership function, $\mathbf{m}$, of each respective feature (row) for each scene class (column). This is opposed to \cite{Boutell_Luo_Shen_Brown_2004,Zhang_2007} which require human intervention in manually annotate the training data as prior information.

In order to learn the 4-tuple fuzzy number, we have chosen to use the histogram representation. As illustrated in Fig. \ref{fig:mf}, it consists of tabular frequencies, shown as adjacent rectangles, erected over discrete intervals (bins), with an area equal to the frequency of the observations in the interval. The height of a rectangle is also equal to the frequency density over the interval, i.e., the frequency divided by the width of the interval. The total area of the histogram is equal to the number of data. 

More specifically, a histogram is a function that counts the number of observations, $n$, that fall into each of the disjoint categories (known as bins), whereas the graph of a histogram is merely one way to represent a histogram. Thus, if we let $N$ be the total number of observations and $B$ be the number of bins, then $N = \sum\limits_{i = 1}^B {{n_i}}$. In the proposed method, for every feature and class label, $\mathbf{x}_{jk}=\{x_i^{jk}\}_{i=1}^N$, we create a histogram in order to obtain the $\mathbf{m}_{jk}$.

We utilize the histogram in representing the occurrence of the training data to the corresponding feature values with an empirical bin width. There is no "best" number of bins, and different bin sizes can reveal different features of the data. Some theoreticians have attempted to determine an optimal number of bins \cite{wand1997data,dalal2005histograms,shimazaki2007method}, but these methods generally make strong assumptions about the shape of distribution. Depending on the actual data distribution and the goals of analysis, different bin number may be appropriate. So an experiment is usually needed for this purpose. Similarly, we utilize (\ref{eq:v}) to find the bin width, $v$. 

\begin{equation}
\left\lceil {v = \frac{{ \wedge x -  \vee x}}{B}} \right\rceil
\label{eq:v}
\end{equation}

\noindent where $\lceil\,\bullet\,\rceil$ indicates the ceiling function and $B=50$ is the total number of bins chosen empirically in our framework. We calculate the occurrence of the training data in each bin and yield a feature vector of $N=\{n_1,n_2,\cdots,n_B\}$. From here, we locate the dominant region, ${\bar \mu }$.

\begin{equation}
{\bar \mu }=\frac{\sum_{i=1}^{B}n_i}{b}
\label{eq:muN}
\end{equation}

\noindent where $b$ denoted the total number of bin which satisfy $n>0$. 

The dominant region (mutually exclusive) is defined as the region where the distribution of training data is higher than $\bar \mu$. We mark this region with membership value equals to 1. By referring to Fig. \ref{fig:mf}, the parameters of $a$ and $b$ of $\mathbf{m}$ can be determined as the lower and upper bound of the area that possess membership value equals to 1. The intersection region (non-mutually exclusive) $a-\alpha$ and $b+\beta$ can be determined as the lower and upper bound of the area that possess membership value equals to 0 respectively. Algorithm 1 summarizes the learning process with a set of training image with $K$ classes. 

\begin{algorithm}
\label{Algo:Modeling}
\caption{\textsc{Learning Framework}}
\begin{algorithmic}
\REQUIRE A training dataset
\STATE {\bf Step 1: Grouping images} Group every image to its respective class label, $\mathbf{I}\to \{ \mathbf{I}^k \}_{k=1}^K$.
\STATE {\bf Step 2: Acquiring the feature values} for all $\mathbf{I}^k$, perform preprocessing to obtain $\mathbf{x}_k$ where $J$ attributes are acquired. Then compute $\mathbf{x}_{jk}=\{x_i^{jk}\}_{i=1}^N$.
\STATE {\bf Step 3: Learning Model}\\
{\bf for all} j such that $1\leq j \leq J$ {\bf do} \\
\quad {\bf for all} k such that $1\leq k \leq K$ {\bf do} \\
\quad\quad Build a histogram of $\mathbf{x}_{jk}$ \\
\quad\quad Compute $\bar \mu$ with (\ref{eq:muN}) \\
\quad\quad Obtain $\mathbf{m}_{jk} = \{a,b,\alpha,\beta\}_{jk}$ based on $\bar \mu$ \\
\quad {\bf end for} \\
{\bf end for} \\
\RETURN $\mathbf{M}$
\end{algorithmic}
\end{algorithm}

\subsection{Inference}
\label{Cla}

Our goal here is to relax the mutually-exclusive assumption on the scene data and classify an unknown scene class into their possibility scene classes and therefore, one scene image can belongs to multiple scene classes. This is unlike the conventional fuzzy inference engine that the de-fuzzification step eventually derives a crisp decision.

Given a testing scene image and its respective feature values $x$, the membership value $\mu$ of feature $j$ belong to class $k$ can be approximated by (\ref{eq:ftset}).

{\begin{equation}
\label{eq:ftset}
\mu_{jk}(x_j) = \left\{ {\begin{array}{*{20}c}
0, & {x_j < a - \alpha} \\
{\alpha ^{ - 1} \left( {x_j - a + \alpha } \right)}, & {a-\alpha \leqslant x_j < a} \\
1, & {a \leqslant x_j \leqslant b} \\
{\beta ^{ - 1} \left( {b + \beta - x_j} \right)}, & {b < x_j \leqslant b+\beta} \\
0, & {x_j > b + \beta } \\
\end{array} } \right.
\end{equation}} 

\noindent where the parameter $a,b,\alpha,$ and $\beta$ are retrieved from $\mathbf{m}_{jk}$ of the learnt FQRC model. We then calculate the product, $P_k$ of membership values of all the attributes for each class, $k$ using (\ref{eq:TP}). Finally, we normalize the $P_k$ and denote as ${r_k}$ (\ref{eq:NTP}),

\begin{equation}
P_k=\prod_{j=1}^{J}\mu _{jk}(x_j)
\label{eq:TP}
\end{equation}

\begin{equation}
{r_k}=\frac{P_k}{\sum P} = \frac{\prod_{j=1}^{J}\mu _{jk}(x_j)}{Z}
\label{eq:NTP}
\end{equation}

\noindent where $Z$ is the normalizer. The intuition to use the product of membership values of all the attributes for each scene class, $P_k$ is to calculate the confident value of each class. This is the core to relate the inference mechanism closer to the principle of human reasoning and relax that scene images are mutually exclusive. If the attribute of a testing data is dominantly belonged to a certain class, $k$ (which means the membership value of that particular attribute, $\mu_{jk} = 1$), and the same for other attributes, at the end of the $P_k$ calculation, testing data belongs to that particular class will be very definite because the product between values 1 is still equal to 1. On the other hand, if the uncertainties for the attributes (membership value of the attribute $\mu_{jk}<1$) are cumulated, the confident value decreases. In mathematical view, the products between values of less than 1 will eventually produce smaller value.

\subsubsection{Summary}
\label{ExpC}

\begin{figure}[htbp]
\centering
\subfigure[Class 1, $\mu=1.0000$]{\includegraphics[scale=0.25]{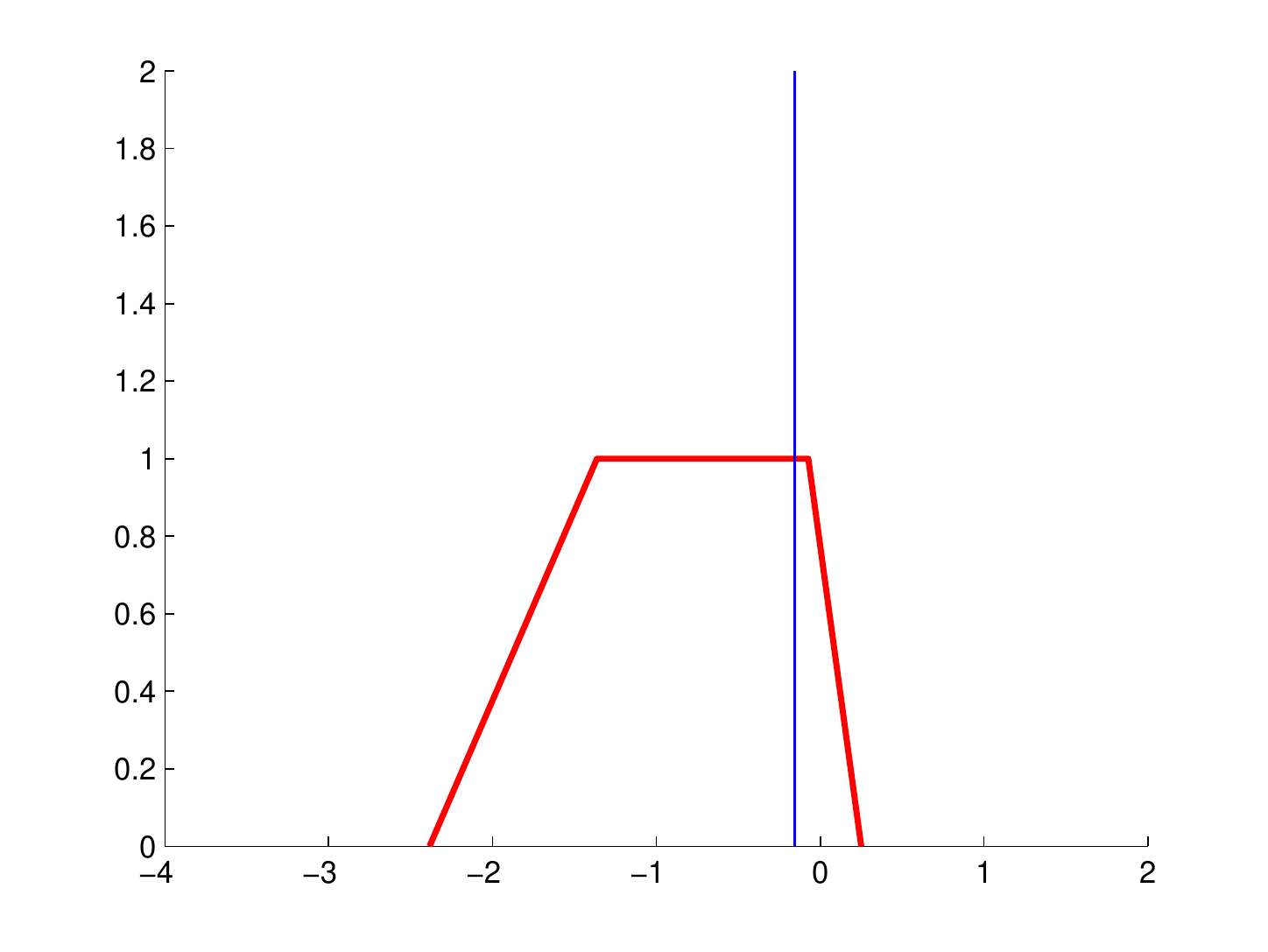}
\label{pic_ex4class2attr_11}}
\subfigure[Class 1, $\mu=1.0000$]{\includegraphics[scale=0.25]{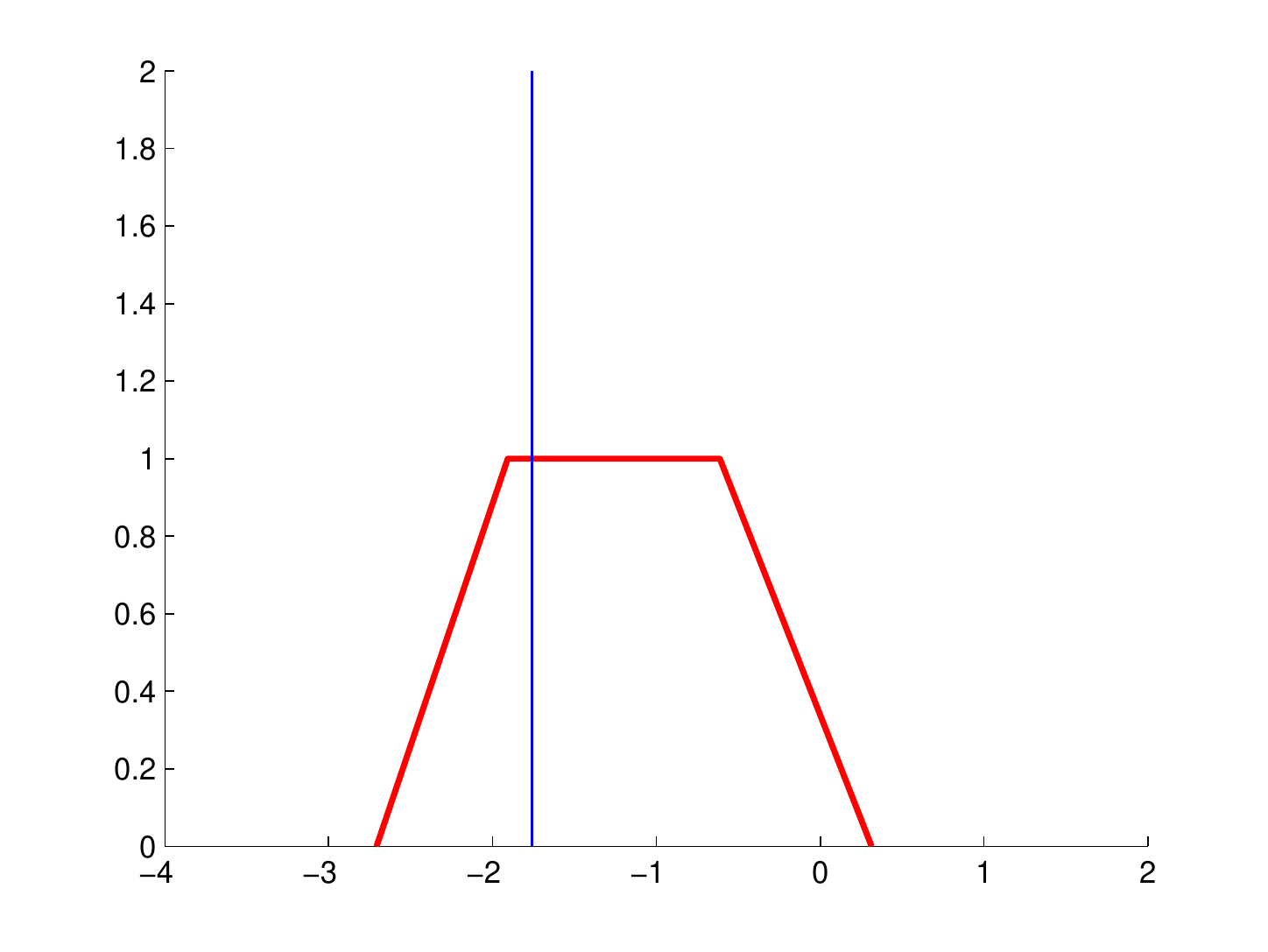}
\label{pic_ex4class2attr_21}}
\subfigure[Class 2, $\mu=0.3046$]{\includegraphics[scale=0.25]{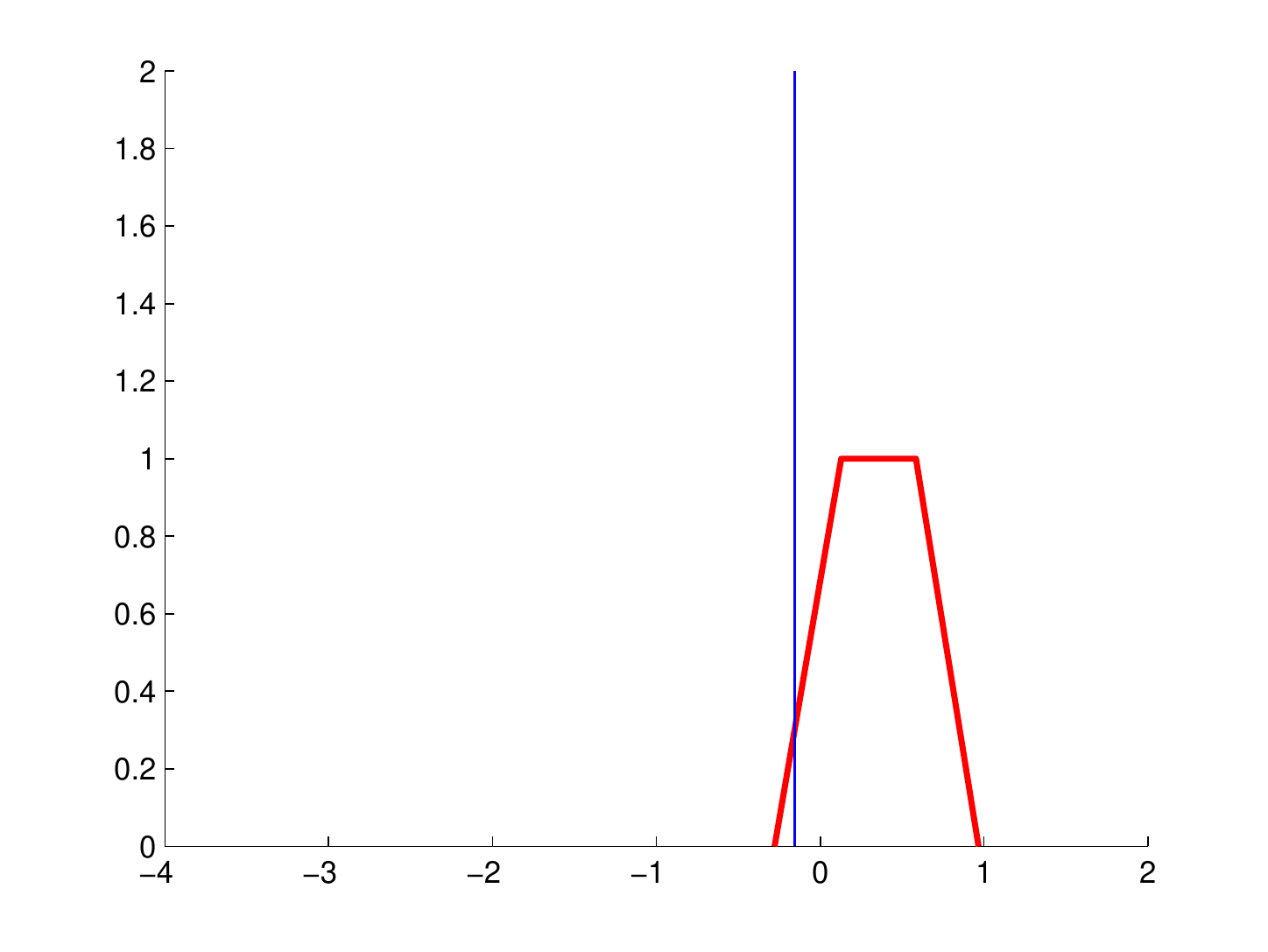}
\label{pic_ex4class2attr_12}}
\subfigure[Class 2, $\mu=0.1558$]{\includegraphics[scale=0.25]{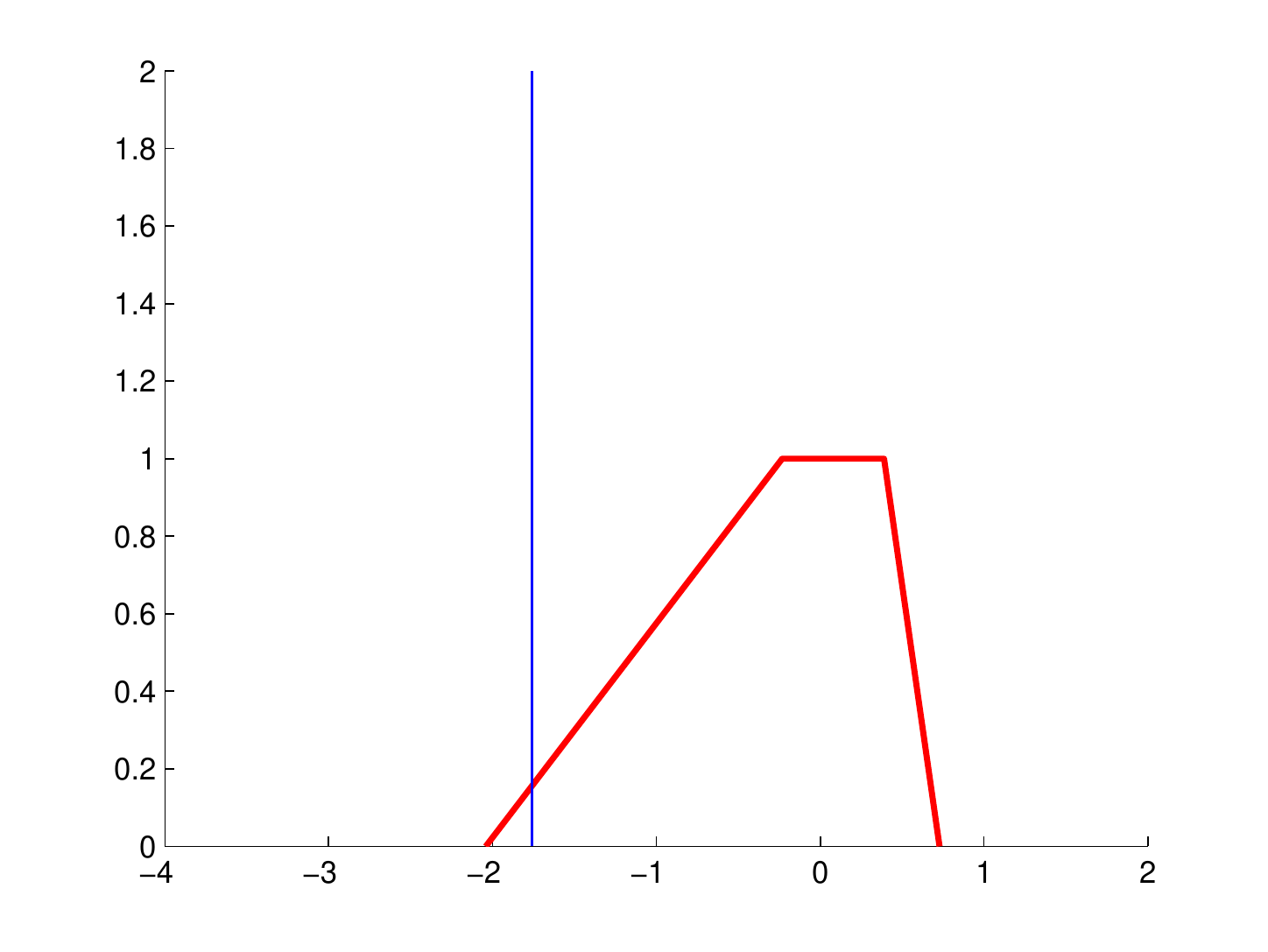}
\label{pic_ex4class2attr_22}}
\subfigure[Class 3, $\mu=0.5406$]{\includegraphics[scale=0.25]{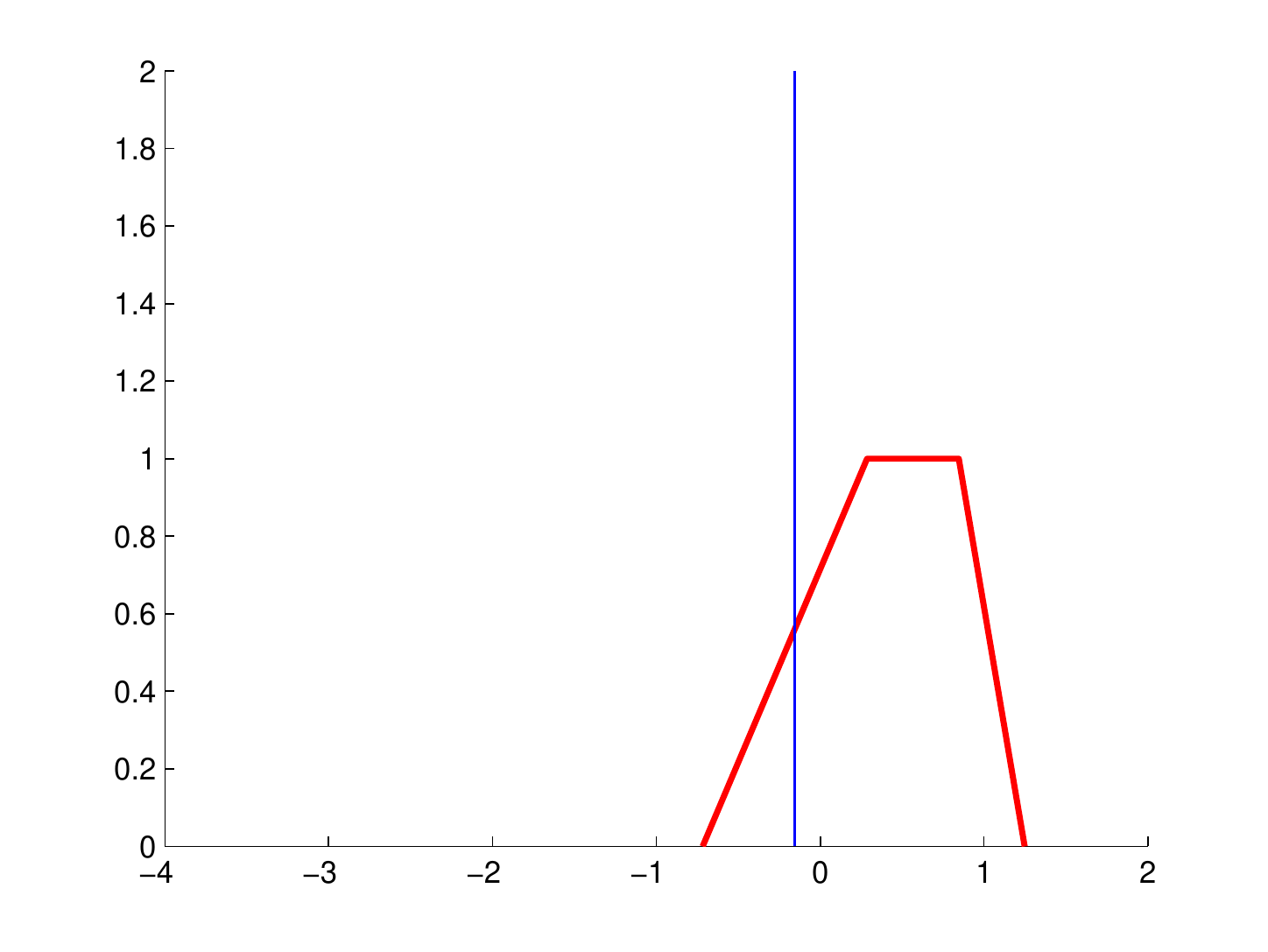}
\label{pic_ex4class2attr_13}}
\subfigure[Class 3, $\mu=0.0000$]{\includegraphics[scale=0.25]{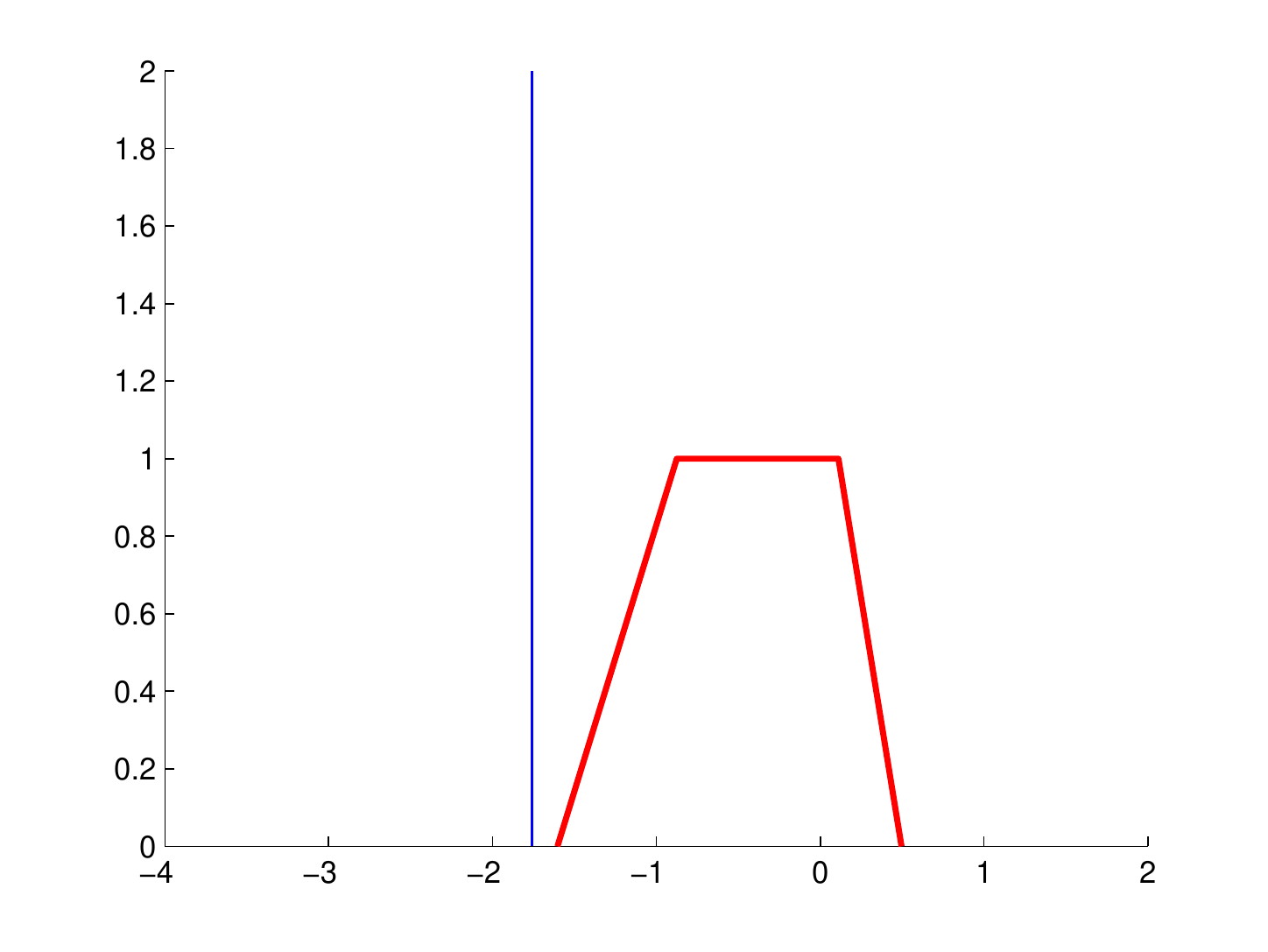}
\label{pic_ex4class2attr_23}}
\subfigure[Class 4, $\mu=0.7508$]{\includegraphics[scale=0.25]{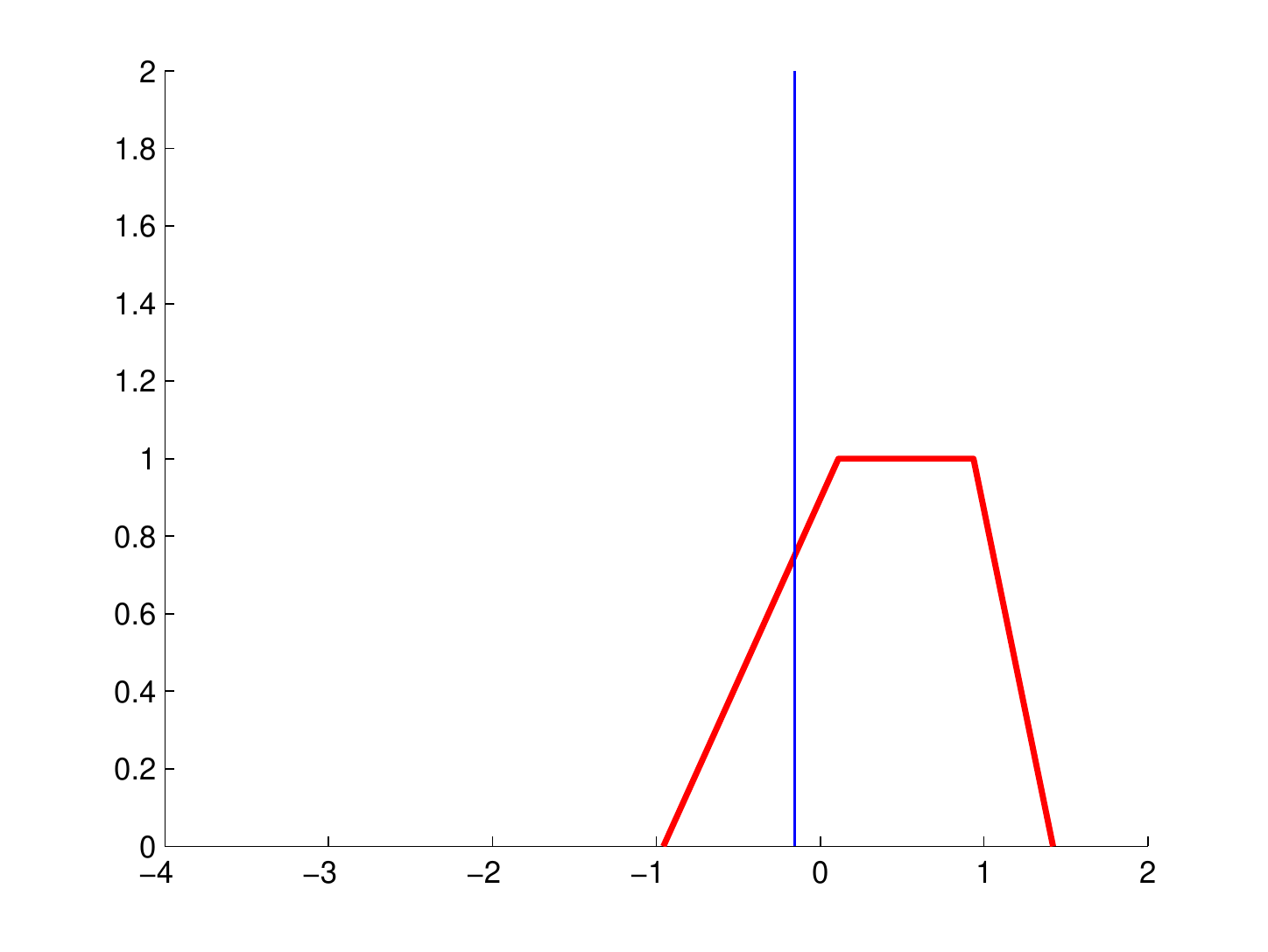}
\label{pic_ex4class2attr_14}}
\subfigure[Class 4, $\mu=1.0000$]{\includegraphics[scale=0.25]{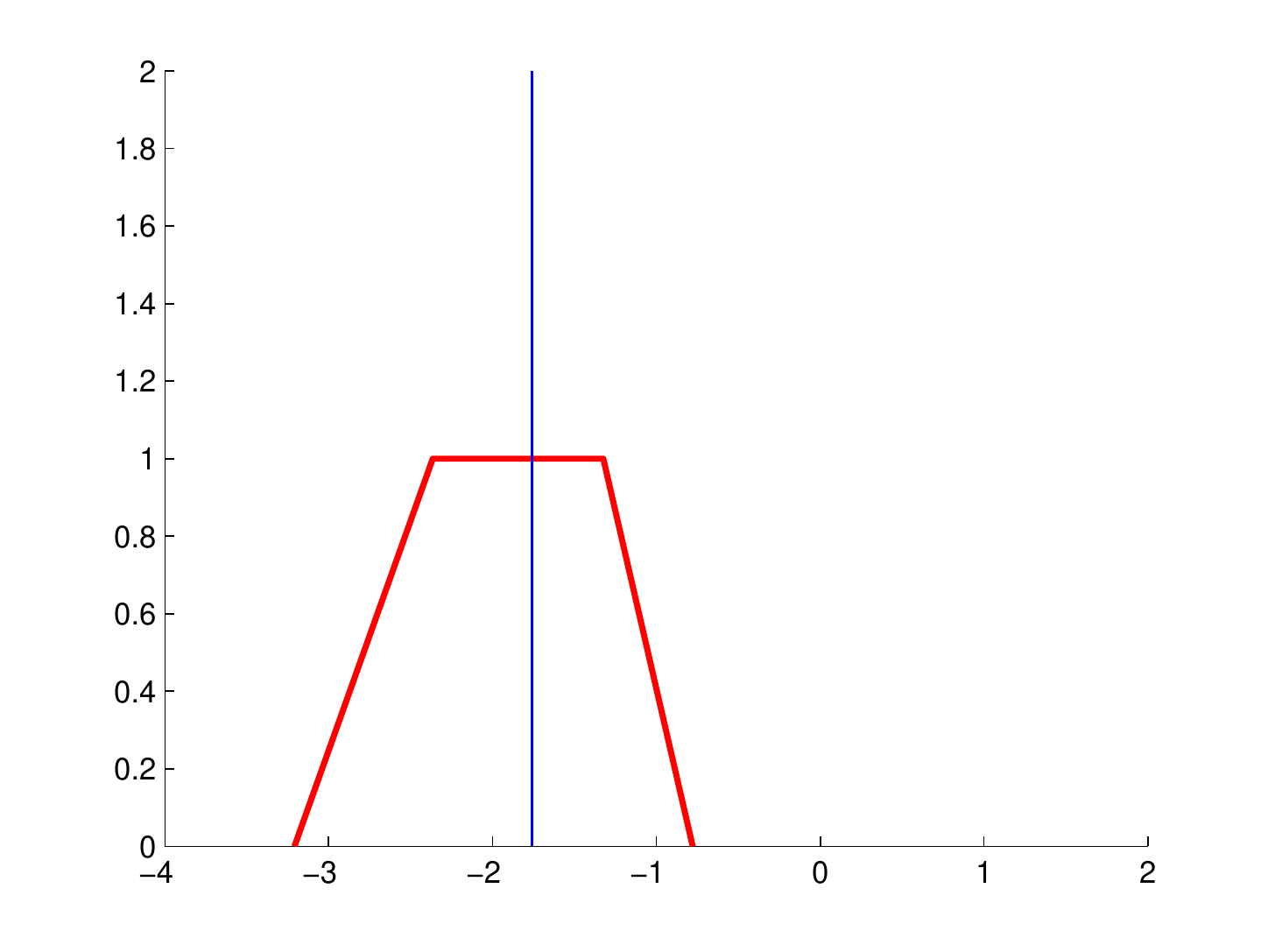}
\label{pic_ex4class2attr_24}}
\caption{The degree of membership, $\mu$, of the attributes (`Natural' on the left, `Open' on the right) for the respective classes. }
\label{fig:attmfs}
\end{figure}

Fig. \ref{fig:attmfs} and \ref{image4class_g} show an example walk-through with a testing image, $s$ and a learnt FQRC model. Let us denote the attributes of the testing image (Fig. \ref{image4class_g}) as $x_1= -0.1545$ and $x_2 = -1.7597$, respectively. For simplicity, we used only 2 attributes in this example but not limited to. By employing the learnt FQRC model (Fig. \ref{fig:attmfs}), we compute $P_k$ as to (\ref{eq:TP}) and $r_k$ as (\ref{eq:NTP}).

To the end, we obtain $r_1 = 0.5561$, $r_2 = 0.0264$, $r_3 = 0.0000$ and $r_4 =
0.4175$, respectively. Each of these values represent that the scene
$s$ has the confident value $r_1$ belongs to ``Insidecity'', $r_2$
belongs to Coast, $r_3$ belongs to ``Opencountry'', and $r_4$ belongs
to Forest where $\sum r = 1$. From human perspective, this result is
reasonable as in the picture, there are characteristics of
``Incidecity'' and ``Forest''. For examples, there are buildings,
vehicles, as well as trees. Therefore, in the inference process, we observed high degree of memberships of the attributes from both classes and thus infer a high value for $r_1$ and $r_4$. While, on the other hand, it possesses almost zero for $r_2$ and zero for $r_3$ because of low or zero value determined from the respective attributes.

As discussed, most state-of-the-art approaches assumed that scene images are mutually-exclusive. Therefore, different strategies to a built a sophisticated binary classifier (inference engine) were proposed in those state-of-the-art approaches. As opposed to these solutions, our work argued that scene images are non-mutually exclusive. Hence, our inference engine contributes in such a way that where ranking interpretation replaces the binary decision. Nevertheless, we provide comprehensive study and stability analysis of our proposed framework in the following section.

\section{Fuzzy membership function and stability analysis}
\label{disOfRQRC}
Before we proceed to the final stage of the FQRC, we provide the intuition of using the 4-tuple membership function in our proposed framework to solve the non-mutually exclusive problem. In addition, the stability analysis of our overall framework is discussed.

\subsection{Fuzzy membership function}
\label{ap:4t}
In this section, we discuss the intuitive idea of using 4-tuple fuzzy membership function in our framework. If we define our loss function as,

\begin{equation}
\ell(f_i(\textbf{x}),y) = \begin{cases}
0 & \text{if } y = \underset{k\in \{1,\ldots,K\}}{\max}\textbf{r}^i(\textbf{x}) \\
1 & \text{otherwise}
\end{cases}
\label{eq:lossfunc}
\end{equation}

\noindent where $\textbf{r}^i(\textbf{x}) =\{r_1^i,\ldots,r_K^i |r_k^i \in [0,1]\}$ is the output of the inference of function $i$, the scalar output $r_k^i$ is defined in (\ref{eq:NTP}) and $\sum_{k=1}^{K}r_k^i=1$. Suppose we have finitely $g$ functions, then, our objective is to find a function $f^*(\textbf{x})$ that minimize the loss function,

\begin{equation}
f^*(\textbf{x}) = \underset{y\in \{1,\ldots,K\}}{\text{arg min }} \sum_{i=1}^{g} \ell(f_i(\textbf{x}),y)
\label{eq:objfunc}
\end{equation}

In order to get the interpretation of (\ref{eq:objfunc}) we will use the concept of maximum entropy. In information theory, the principle of maximum entropy is to minimize the amount of prior information built into the distribution. More specifically, the structure of maximum entropy problem is to find a probability assignment (or membership function $\mu_{jk} \in [0,1]$) which avoid bias agreeing with any given information. In this case, while looking at
(\ref{eq:objfunc}), the membership function $\mu_{jk}$ captures such prior information. Inspired by Miyamoto and Umayahara \cite{miyamoto1998fuzzy}, we utilize the maximum entropy to get the interpretation of 4-tuple. For simplicity we omit $i$, and the objective of maximum entropy,

\begin{equation}
\max -\sum_j \sum_k \mu_{jk} \log \mu_{jk}
\end{equation}

Subject to the constraint $\sum_k \frac{\prod_j \mu_{jk}}{Z}=1$ and $f^*(\textbf{x})=c$, where $c$ is a constant. Then using Lagrange multipliers,

\begin{equation}
\begin{aligned}
\mathcal{J} = & -\sum_j \sum_k \mu_{jk} \log \mu_{jk} + \lambda_1 \left(1 - \sum_k \frac{\prod_j  \mu_{jk}}{Z} \right) \\
 &+ \lambda_2 (c-f^*(\textbf{x}))
 \end{aligned}
\end{equation}

For simplicity, we treat $\mu_{jk}$ as a fixed length vector since we assume $\textbf{x}$ is discrete, then we have,

\begin{equation}
\frac{\partial \mathcal{J}}{\partial \mu_{jk}} = -1 - \log \mu_{jk} - \frac{\lambda_1}{Z} - \lambda_2 \frac{\partial f^*(\textbf{x})}{\partial \mu_{jk}}
\end{equation}

\noindent Setting $\frac{\partial \mathcal{J}}{\partial \mu_{jk}} = 0$ and get $\mu_{jk}$ yields,

\begin{equation}
\mu_{jk} = \exp \left( -\left( 1+\frac{\lambda_1}{Z} + \lambda_2 \frac{\partial f^*(\textbf{x})}{\partial \mu_{jk}} \right) \right)
\end{equation}

\noindent Actually this result is similar when we minimize or maximize the objective function of,

\begin{equation}
\text{min/max} -\sum_j \sum_k \mu_{jk} \log \mu_{jk} -\lambda_2 f^*(\textbf{x})
\end{equation}

With subject to the constraint $\sum_k \frac{\prod_j \mu_{jk}}{Z}=1$. After taking min-max sign change and make the constant $\lambda=1/\lambda_2$ for brevity, we get the following objective,

\begin{equation}
\begin{aligned}
& \min f^*(\textbf{x}) - \lambda \sum_j \sum_k \mu_{jk} \log \mu_{jk} \\
& \text{subject to} \sum_k \frac{\prod_j \mu_{jk}}{Z}=1
\end{aligned}
\label{eq:obj_regularize}
\end{equation}

If we compare (\ref{eq:obj_regularize}) with the formula of a classifier with regularization, $f + \lambda \mathcal{R}$, the 4-tuple membership function implicitly models the regularization. In details, the 4-tuple membership function with $\mu_{jk}=1$ (mutually exclusive part) models the classifier while the transition of membership function $[0,1]$ (non-mutually exclusive part) implicitly models the regularization.

\subsection{Stability Analysis}
\label{ap:sa}
In this section, we discuss the robustness of the proposed framework in terms of stability analysis. In particular, the concept of stability brought by Bousquet and Elisseeff \cite{bousquet2002stability} is employed as it gives guarantee of a ``good" learning machine by deriving generalization error bounds. As a matter of fact, the error bounds are derived from the stability. More specifically, the stability measures how much the output will change for small changes in the training data. One said an algorithm that is stable, whose output will not depend on a single sample, tends to have generalization error that is close to the empirical error bounded by a constant. We define stability as follows,

\begin{defn} [Stability]
Let $(\textbf{x}_i,y_i)=z_i (\in \mathcal{Z})$ be a sample from a set of samples $\mathcal{Z}$ and $(\mathcal{Z}\backslash z_i) \cup z'_i$ be a set of samples after replacing a sample $z_i$ with a new sample $z'_i$ which is independent from $\mathcal{Z}$. A function $f:\mathcal{Z}^N \to \mathbb{R}^N$ has stability with respect to the loss function $\ell$ by a constant $\beta$, such that

\textbf{\begin{equation}
\begin{aligned}
\forall \mathcal{Z} \in (\mathcal{X} & \times \mathcal{Y})^N, \forall i \in \{1,\ldots,N\} : \\ 
& \mathbb{E}_\mathcal{Z} | \ell(f_\mathcal{Z},\cdot) - \ell(f_{(\mathcal{Z}\backslash z_i) \cup z'_i},\cdot) | \leq \beta
\label{eq:stability}
\end{aligned}
\end{equation}}

We call $f$ stable. Similarly, for large classes of functions if for all $f \in \mathcal{F}$ satisfy condition (\ref{eq:stability}), then $\mathcal{F}$ is stable as well. The constant $\beta$ should be on the order of $O(\frac{1}{N})$ \cite{bousquet2002stability}.
\label{defn:stability}
\end{defn}

If we define the empirical risk as $R_{emp}=\frac{1}{N}\sum_{z_i \in \mathcal{Z}}\ell (f(\textbf{x}),y)$, for simplicity we will denote $R=\frac{1}{N}\sum_{z_i \in \mathcal{Z}}\ell_{z_i}$, and let $R_{\mathcal{Z} \cup z'}$ be the empirical risk after adding a sample $z'$.

\begin{equation}
\begin{aligned}
R_{\mathcal{Z} \cup z'} &= \frac{1}{N+1} \sum_{z_i\in (\mathcal{Z} \cup z')} \ell_{z_i} \\
&= \frac{1}{N+1} \left( \sum_{z_i\in \mathcal{Z}} \ell_{z_i} + \ell_{z'} \right) 
= \frac{N}{N+1} \left( R + \frac{1}{N}\ell_{z'} \right)
\end{aligned}
\end{equation}

\noindent Similarly, we get the risk after removing a sample $z''$, that is $R_{\mathcal{Z} \backslash z''}$,

\begin{equation}
R_{\mathcal{Z} \backslash z''} = \frac{N}{N-1} \left( R + \frac{1}{N}\ell_{z''} \right)
\end{equation}

\noindent Then the risk after we replace a sample,
$R_{(\mathcal{Z}\backslash z'') \cup z'}$, can be denoted as follows,

\begin{equation}
\begin{aligned}
R_{(\mathcal{Z}\backslash z'') \cup z'} &= \frac{N-1}{N+1-1} \left[ R_{\mathcal{Z} \backslash z''} + \frac{1}{N-1} \ell_{z'}\right] \\
&= \frac{N-1}{N} \left[ \frac{N}{N-1}\left( R-\frac{1}{N}\ell_{z''} \right) + \frac{1}{N-1} \ell_{z'} \right] \\
&= R - \frac{1}{N} \ell_{z''} + \frac{1}{N} \ell_{z'}
\end{aligned}
\end{equation}

\noindent Using the triangle inequality and by noting (\ref{eq:lossfunc}) that $\ell_{z'},\ell_{z''} \leq 1$,

\begin{equation}
| R_{(\mathcal{Z}\backslash z'') \cup z'} - R | \leq \beta_b \leq \frac{2}{N}
\label{eq:risk_beta}
\end{equation}

\noindent where $\beta_b$ is the stability of the underlying binary
classifier. In order to get the stability of our method, the loss
function (\ref{eq:lossfunc}) must be $\sigma$-admissible for any
$\sigma$ (it is also need to be convex)
\cite{miyamoto1998fuzzy,bousquet2002stability,rifkin2002everything}. Let's
define the  total loss as $\ell_{tot}(\textbf{x},y)=\sum_{i=1}^{g}
\ell(f_i(\textbf{x},y))$, then our new loss function (parameterized
over $\gamma > 0$) can be defined as,

\begin{equation}
{\small
\begin{gathered}
  {\ell _\gamma }(f(\textbf{x}),y) =  \hfill \\
  \left\{ {\begin{array}{*{20}{l}}
  0&{{\text{ if }}{\ell _{tot}}(\textbf{x},y) < {{\wedge }_{k \ne y}}{\ell _{tot}}(\textbf{x},k)} \\ 
  {\frac{{{\ell _{tot}}(\textbf{x},y) - {{\wedge }_{k \neq y}}{\ell _{tot}}(\textbf{x},k)}}{\gamma }}&{{\text{if 0}} \leqslant {\ell _{tot}}(\textbf{x},y) - {{\wedge }_{k \neq y}}{\ell _{tot}}(\textbf{x},k) \leqslant \gamma } \\ 
  1&{{\text{ if }}{\ell _{tot}}(\textbf{x},y) - {{\wedge }_{k \ne y}}{\ell _{tot}}(\textbf{x},k) > \gamma } 
\end{array}} \right. \hfill \\ 
\end{gathered}}
\end{equation}

\noindent where $k, y \in \{1,\ldots,K\}$ are class labels and
$\wedge = \min$. It is clear that for any $\gamma > 0$ then
$\ell_\gamma (f(\textbf{x}),y) \geq \ell(f(\textbf{x},y))$ and it is
$\sigma$-admissible (in fact, $1/\gamma$-Lipschitz with respect to its
first argument). In addition, \cite{bousquet2002stability} has shown
that for an algorithm with regularization, $f+\lambda \mathcal{R}$, it
contributes to the bounded constant by
$\frac{1}{\lambda}\beta_b$. Combining $\sigma-$admissible and
regularization, $\frac{1}{\lambda}\beta_b$, (\ref{eq:risk_beta}) becomes,

\begin{equation}
| R_{(\mathcal{Z}\backslash z'') \cup z'} - R | \leq \frac{2}{\gamma \lambda N}
\label{eq:risk_beta1}
\end{equation}

Indeed, this result satisfies the definition of stability as stated in Definition \ref{defn:stability}. More specifically, when we replace a single sample, the loss function $\ell_\gamma$ will change by at most
$\frac{2}{\gamma \lambda N}$, meaning that $\ell_{tot}(\textbf{x},y)$ might increase while $\min_{r\neq y}\ell_{tot}(\textbf{x},r)$ might decrease by $\beta_b$. Thus, a naive bound on the stability of the multiclass system is $\frac{2K}{\gamma \lambda N}$ \cite{rifkin2002everything}.

In order to get the generalization error bound, we use the Bousquet and Elisseeff \cite{bousquet2002stability} theorem.

\begin{thm}[Bousquet and Elisseeff \cite{bousquet2002stability}]
A $\beta$-stable function $f$ satisfying $0 \leq \ell(f_S,z) \leq M$ for all training sets $S$ and for all $z\in \mathcal{Z}$. For all $\epsilon > 0$ and all $N \geq 0$,

\begin{equation}
P\{R - R_{emp} > \epsilon + 2\beta\} \leq \exp \left(  -\frac{2N\epsilon^2}{(4N\beta + M)^2} \right)
\end{equation}

\noindent It gives the following bounds with probability at least $1-\delta$,

\begin{equation}
R \leq R_{emp} + 2\beta + (4N\beta + M)\sqrt{\frac{\ln 1/\delta}{2N}}
\end{equation}

\end{thm}

By substituting $\beta = \frac{2K}{\gamma \lambda N}$ and note that the loss function has maximum value $M = 1$, we get the following bound on multiclass classification for our proposed method,

\begin{equation}
R \leq R_{emp} + \frac{4K}{\gamma \lambda N} + (\frac{8K}{\gamma \lambda} + 1)\sqrt{\frac{\ln 1/\delta}{2N}}
\end{equation}

We summarize here, our proposed method has shown to possess stability since the error is bounded by a constant $\beta = \frac{2K}{\gamma \lambda N}$. Moreover, our framework implicitly models regularization thereby it improves the stability (indicated by $\frac{\beta}{\lambda}$) and provides generalization error bound.

\section{Ranking Interpretation}
\label{IR}

Ranking system is a very common yet important information representation technique in many applications, and recently it has received more attention on applying it in inferring the output of the computer vision algorithm \cite{Parikh_Grauman_2011,Kovashka_Parikh_Grauman_2012}. From the general definition, a ranking is a relationship between a set of items such that, for any two items, the first is either ``ranked higher than'', ``ranked lower than'' or ``ranked equal to'' the second. 

In the previous section, we obtained the normalized product, ${r_k}$ as our final output. However, these values do not provide us any meaningful information to understand the scene images. In order to interpret the results into a more useful manner, we introduce a ranking framework as shown in Table \ref{table:Symb} which acts similar to a decoder to decode our results into a ranking manner. With step by step explanation for our ranking interpretation,

\begin{enumerate}  
  \item Obtain the maximum $r$ value, $\vee {r_k}$.
  \item Discard the scene class with ${r_k} = 0$ by marking the class as $\times$, which mean definitely not.
  \item Compute the difference value, $r_{diff}$ between $\vee {r_k}$ and ${r_k}$ (all involved $r$ values) and apply symbolic representation with the predefined threshold as in Table \ref{table:Symb}.
\end{enumerate}

\begin{figure*}[htbp]
\centering
\includegraphics[height=0.3\linewidth, width=0.85\linewidth]{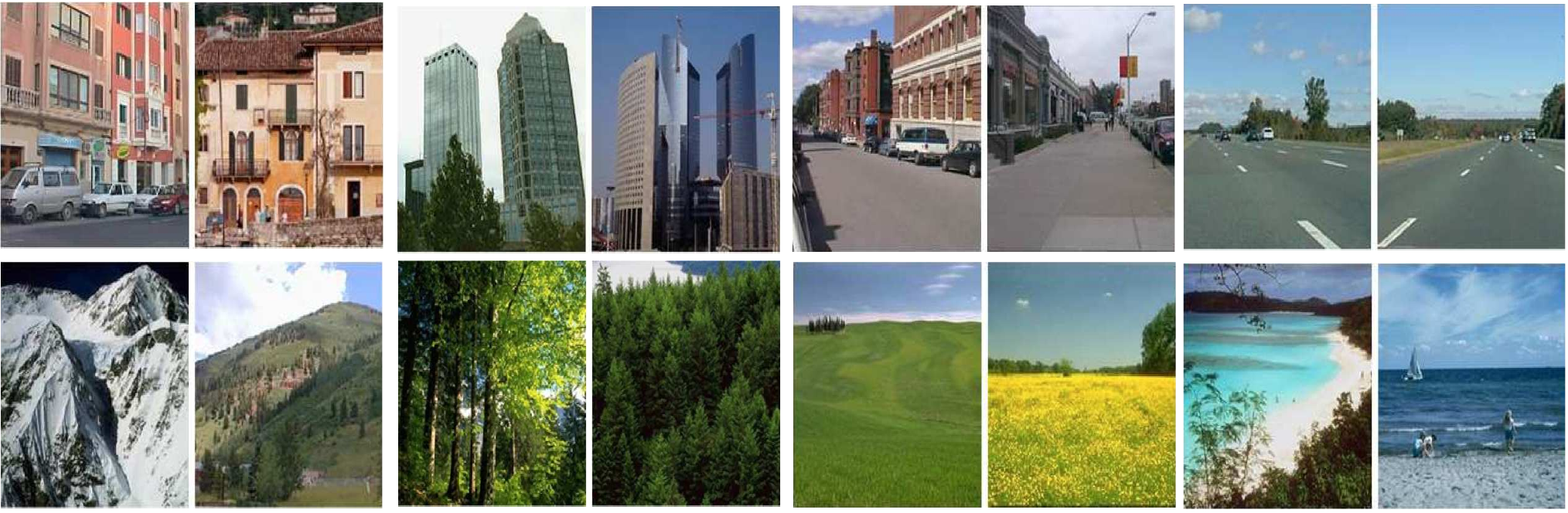}
\caption{Examples of the OSR Dataset. The dataset contains 2688 color scene images, 256x256 pixels from a total of 8 outdoor scene classes. These images are best view in colour.}
\label{fig:ExampleOfScene}
\end{figure*}

Note that, the parameter design in our ranking scheme is depend on the design that fits to the task on hand. In our work, we apply four levels of ranking interpretation with equally divided interval value to fit the different levels of ranking with exception to ``Equal to'' and ``Definitely not''. The four levels of ranking are ``Equal to'', ``Higher than'', ``Much higher than'', and ``Definitely not''. The value of $r_{diff} = [0 \ 1]$ which is the difference between $\vee {r_k}$ with $r_k$ is used to determine the level of particular image compare to the other scene images. Refer to Table \ref{table:Symb}, we define ``Equal to'' as the case when $r_{diff} = 0$, ``Definitely not'' when the value of $r$ for the class involved is equal to 0 (${r_k} = 0$), and the left out is ``Higher than'' and ``Much higher than''. For these two levels, we apply equally divided interval value of the maximum boundary of
$r_{diff}$ which is $0.5$ to partition each level. However, as mentioned above, we are not limiting to this setting as it may varies across the different system designs.

\begin{table}[htbp]
	\centering
		\caption{Symbolic representation of ranking}
		\label{table:Symb}
                {\renewcommand{\arraystretch}{1.5}
		\begin{tabular}{|c||c|c|}		
		\hline	 	
  		Threshold value 								& Symbol 		& Description 	\\	\hline \hline
			0 															& $\equiv$	& Equal To					\\ \hline
			$0 < r_{diff} \leqslant 0.5$ 		& $>$ 			& Higher than 			\\ \hline
			$0.5 < r_{diff} \leqslant 1.0$ 	& $\gg$ 	 	& Much higher than	\\ \hline	\hline \hline
                       ${r_k} = 0$                          & $\times$	 & Definitely not			\\ \hline	
                        
		\end{tabular}}
\end{table}

\begin{table}[htbp]
\centering
{\scriptsize
		\caption{Example of ranking result}
		\label{table:res}
                {\renewcommand{\arraystretch}{1.5}
		\begin{tabular}{|c|c|c|c|}		
		\hline	 		  		
		Scene Class	& ${r_k}$                         & $r_{diff}$	                                                              & Symbol	\\ \hline \hline
		Class 1 	        & 0.5561					& Not applicable as this is $\vee {r_k}$    & Not applicable 			\\ \hline
		Class 2  	        & 0.0264					& 0.3911					                                       & $>$			\\ \hline
                Class 3 	        & 0.0000					& Discarded as ${r_3} = 0$ 					& $\times$ 			\\ \hline
		Class 4 	        & 0.4175					& 0.1386				                                               & $>$			\\ \hline \hline
		Final Result			& \multicolumn{3}{|c|}{Class 1$>$ Class 4$>$ Class 2, but $\times$ Class 3}	\\
                                    			& \multicolumn{3}{|l|}{Scene image has the highest possibility as Class 1,} \\
                                                      & \multicolumn{3}{|l|}{follow by Class 4 and Class 2 respectively.}\\
                                                      & \multicolumn{3}{|l|}{But scene image is definitely not belong to Class 3.}	\\ \hline
		\end{tabular}}}
\end{table}

By applying such method, we can represent a ranking in a symbolic
representation. Using the same example from Section \ref{ExpC}, we
obtain the result as in Table \ref{table:res}. The confident level of
image $s$ belongs to ``Insidecity'' is \textbf{\textit{higher}} than
``Forest'' and ``Coast'', the confident level of image $s$ belongs to ``Forest'' is \textbf{\textit{higher}} than ``Coast'', but image $s$ is definitely not belonged to ``Opencountry''. So at the end of this interpretation method, we are able to obtain the ranking position of every possible classes.

\section{Experiments}
\label{ExpSet}

We tested our approach with two public scene image datasets - the
Outdoor Scene Recognition (OSR) dataset \cite{Oliva_Torralba_2001} and
the Multi-Label Scene (MLS) dataset
\cite{Boutell_Luo_Shen_Brown_2004,Zhang_2007}. The OSR dataset
contains 2688 colour scene images, 256x256 pixels from a total of 8
outdoor scene classes (`Tallbuilding, T', `Insidecity, I', `Street,
S', `Highway, H', `Coast, C', `Opencounty, O', `Mountain, M' and
`Forest, F'). Fig. \ref{fig:ExampleOfScene} illustrates example of the
OSR dataset and is publicly available\footnote{http://people.csail.mit.edu/torralba/code/spatialenvelope}. In the meantime, MLS dataset contains a total of 2407 scene images with 15 (6 base + 9 multi-label) classes. According to \cite{Boutell_Luo_Shen_Brown_2004,Zhang_2007}, the multi-label data in the MLS dataset were manually annotated by three human observers as part of the pre-requirement during the training stage.

In the feature extraction stage for the OSR dataset, we have employed 6 different attributes \cite{Parikh_Grauman_2011} to represent the scene images. The 6 attributes are natural, open, perspective, large objects, diagonal plane and close-depth. Note that, this work is not constrained to these representations. An alternative representation such as other feature extraction methods can be employed as the front-end. Since our focus in this study is the introduction of a fuzzy qualitative approach to perform scene classification, any existing feature representation for images can be employed as the input to our model. In the meantime, for MLS dataset, we employed the feature vector, $\mathbb{R}^{294}$ as to \cite{Boutell_Luo_Shen_Brown_2004,Zhang_2007}. Finally, for learning the model, in OSR dataset, we practiced the `leave-one-out' method and performed classification of each testing image by using the trained model obtained from the rest of the training data. While for the MLS dataset, we followed the distribution of training and testing data in the classification pipeline. All implementations and experiments were conducted in the MATLAB environment. 

Overall, our experiment is divided into five sections (Section \ref{Exp1} to \ref{Exp4}) where each of them is testing on different perspectives. We set the bin number, $B$ of the histogram as 50 and the threshold for the level of ranking interpretation as to Table \ref{table:Symb}.

\subsection{Scene Images are non-Mutually Exclusive}
\label{Exp1}

Psychological and metaphysical \cite{Forguson} proved that there is an influence of human factors (background, experience, age, etc.) in decision making. In this experiment, we would like to show that the research in scene understanding falls within this category and scene images are indeed non-mutually exclusive. For this purpose, an online survey was created with a fair number of scene images, randomly chosen from the OSR dataset. The online survey was run for a month and participated by a group of people in the range of 12 to 60 years old from different backgrounds and countries. Their task is to select a class that best reflects the given scene accordingly without prior knowledge of what the ground truth is.

\begin{figure*}[htbp]
\centering
\includegraphics[height=0.5\linewidth, width=0.9\linewidth]{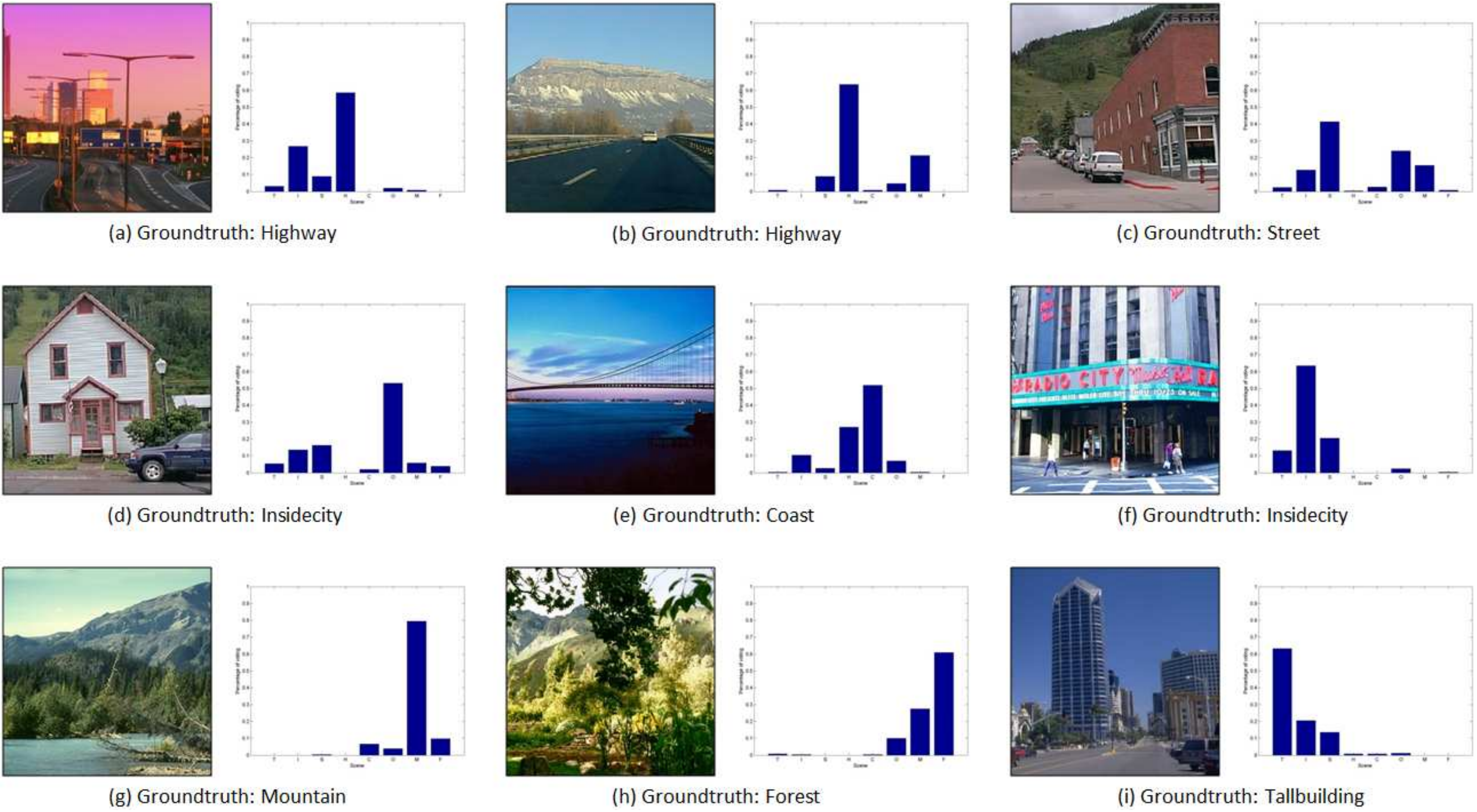}
\caption[]{Examples of the online survey results. We validate that scene images are indeed non-mutually exclusive (from left to right, the bar on each histogram represents the distribution of `Tallbuilding, T', `Insidecity, I', `Street, S', `Highway, H', `Coast, C', `Opencounty, O', `Mountain, M' and `Forest, F' accordingly).}
\label{fig:expSurveyResult}
\end{figure*}

We show some examples of the results from the online survey in Fig. \ref{fig:expSurveyResult}. For a complete result, interested reader is encouraged to look at this website\footnote{http://web.fsktm.um.edu.my/$\sim$cschan/project2.htm}. From here, we can clearly notice that there is a variation of an answer (scene class) for each scene image. For instance, in Fig. \ref{fig:expSurveyResult}(a), although the favorite selection is `Highway' class, the second choice which is `Insidecity' class still occupies noticeable distribution as well. From a qualitative point of view, this observation is valid as the scene image comprises of a lot of buildings that form the city view. Similar to Fig. \ref{fig:expSurveyResult}(h) where the dominant choice is `Forest' class while the second choice of ``Mountain'' class is still valid.   
 
Nevertheless, we should not overpass the minority choices. For
example, in Fig. \ref{fig:expSurveyResult}g, the dominant selection is
a `Mountain' class. However, there are minority who selected `Coast',
`Opencountry' and `Forest', respectively. Even though these choices
are minority, the selections are still valid as we could notice
similar appearance between those choices. Unfortunately, there are
some outliers as depicted in Fig. \ref{fig:expSurveyResult}(a) and
\ref{fig:expSurveyResult}(f) which could be easily eliminated with the $\alpha$-cut that will be explained later.

Besides that, one could observe that the best result from the
histogram of Fig. \ref{fig:expSurveyResult}(a,b,c,e,f,g,h,i) agrees
with the ground truth except for the case in
\ref{fig:expSurveyResult}(c) where the best result is different from
the ground truth. In particular, the image seems to be `Opencountry'
more than `Insidecity'. This is very interesting results to show that
human are bias in identifying a scene image. As a summary, we had
shown that assuming scene images are mutually exclusive and simplify
the classification problem (uncertainty, complexity, volatility and ambiguity) to a binary classification task is impractical as it does not reflect how human reasoning is performed in reality.  

\subsection{Effectiveness of FQRC}

This is to show the correctness of our classifier in handling non-mutually exclusive data and the inconsistency of human decision making. We denote $Y_d$ as the set of result value for scenery image $d$ from the survey and $W_d$ be the set of predicted label from the FQRC. The results are compared in the following aspects:

\subsubsection {Qualitative Observation}

\begin{figure*}[htbp]
\centering
\subfigure[Scene image 1]{\includegraphics[scale=0.5]{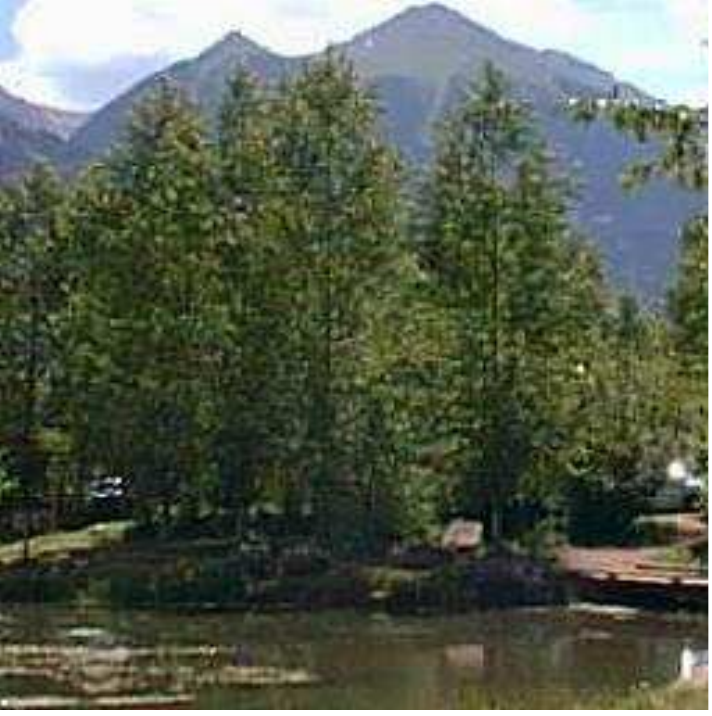}
\label{pic_obs1}}
\subfigure[Result of online survey, $Y_d$]{\includegraphics[scale=0.35]{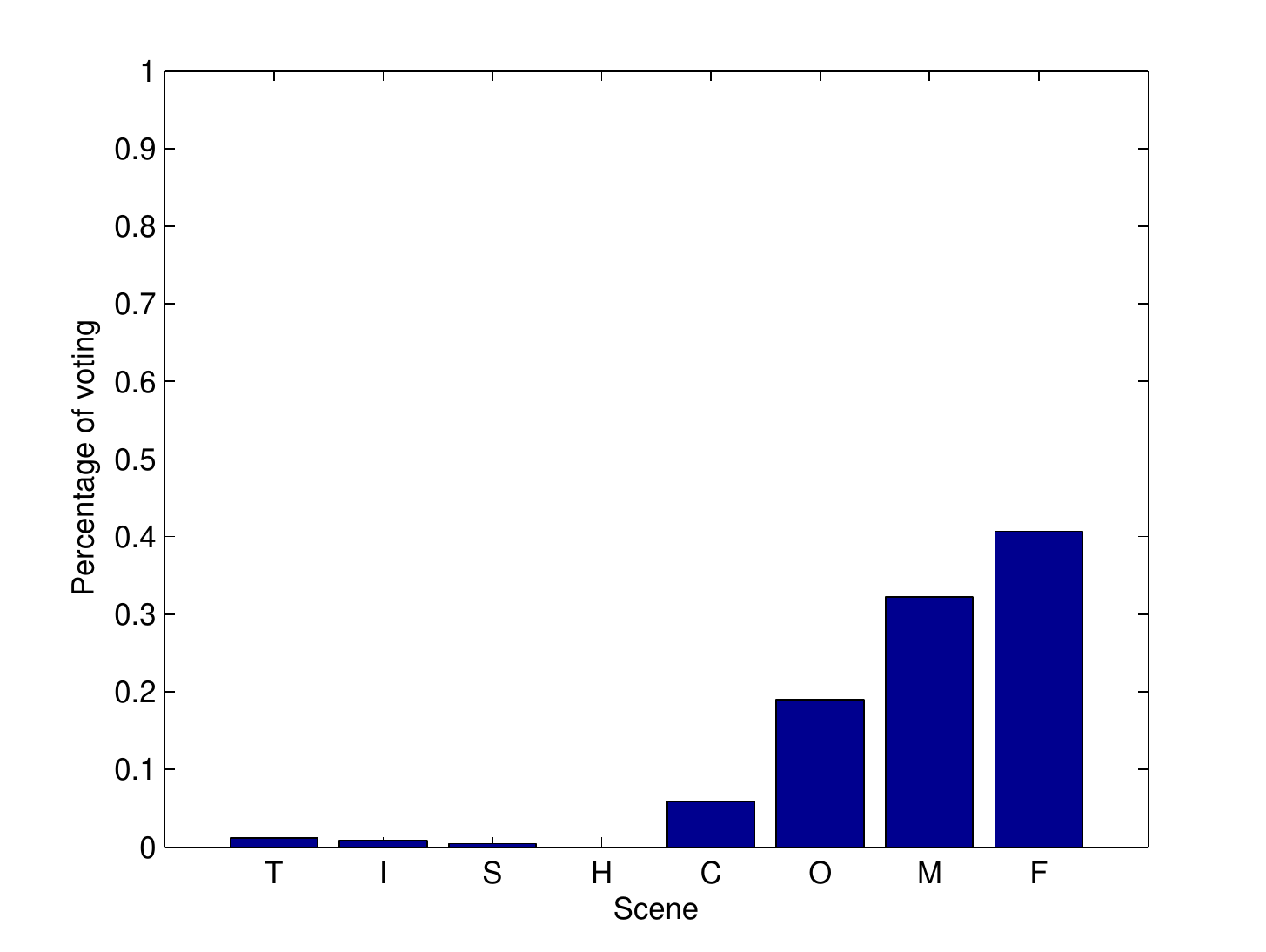}
\label{pic_obsSurvey1}}
\subfigure[Result of FQRC, $W_d$]{\includegraphics[scale=0.35]{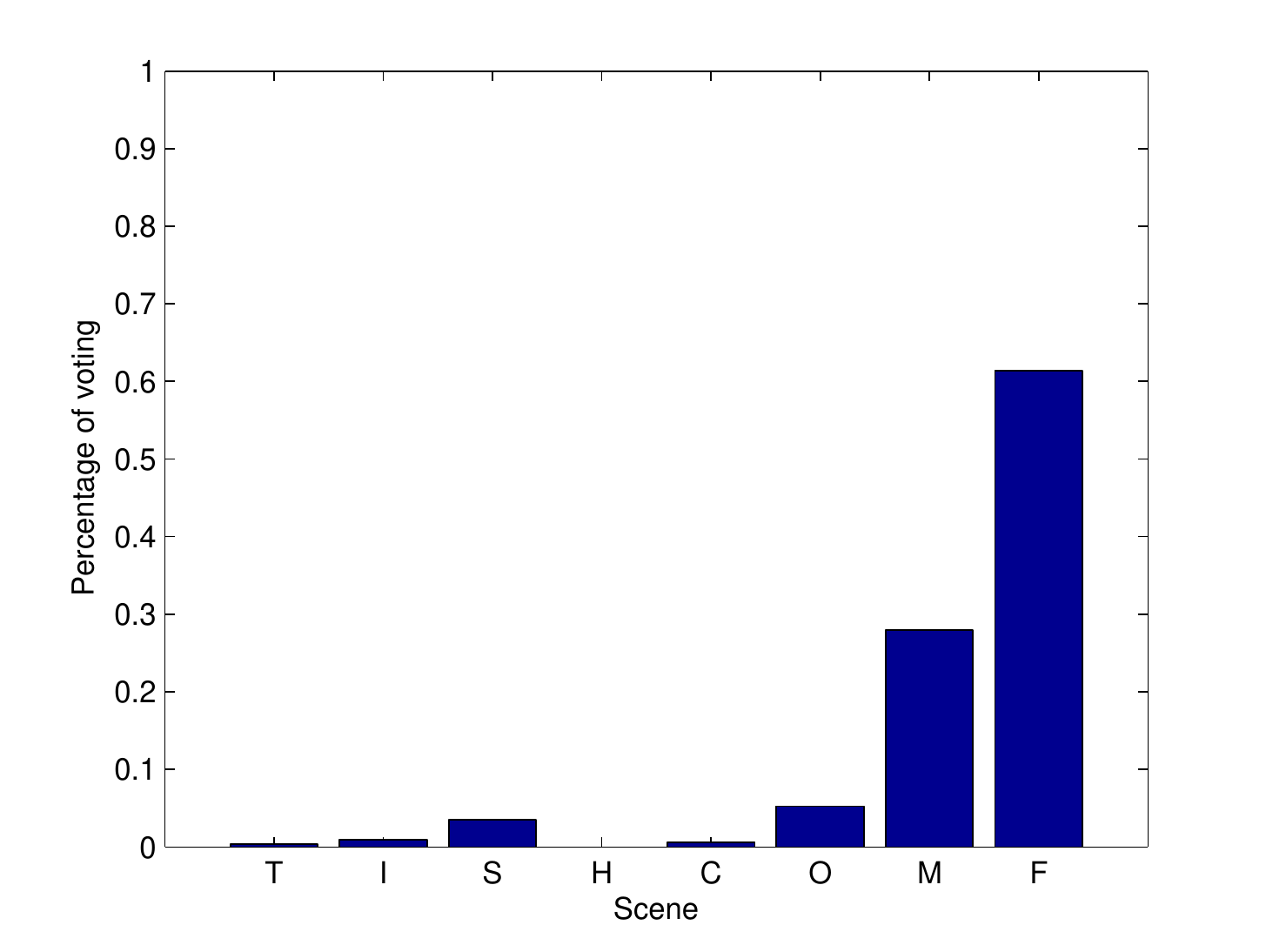}
\label{pic_obsComp1}}
\subfigure[Scene image 2]{\includegraphics[scale=0.5]{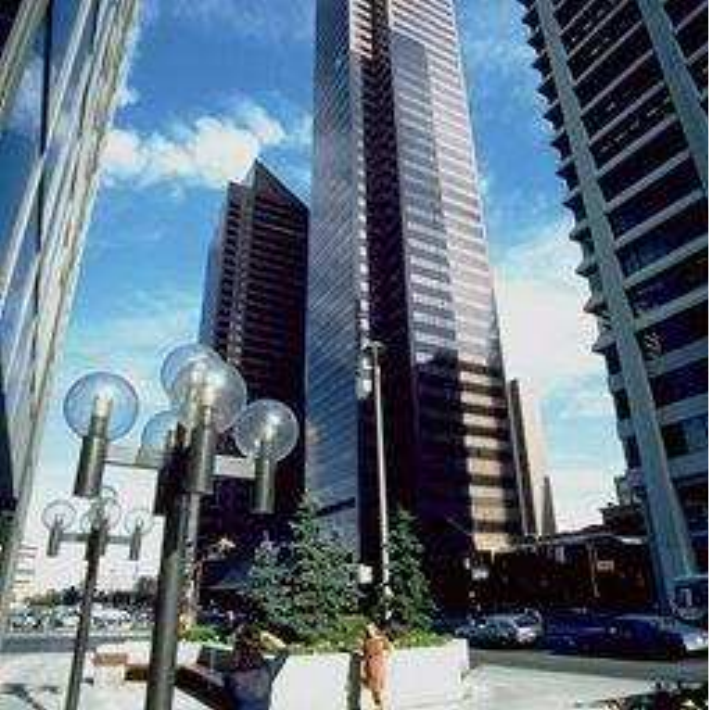}
\label{pic_obs2}}
\subfigure[Result of online survey, $Y_d$]{\includegraphics[scale=0.35]{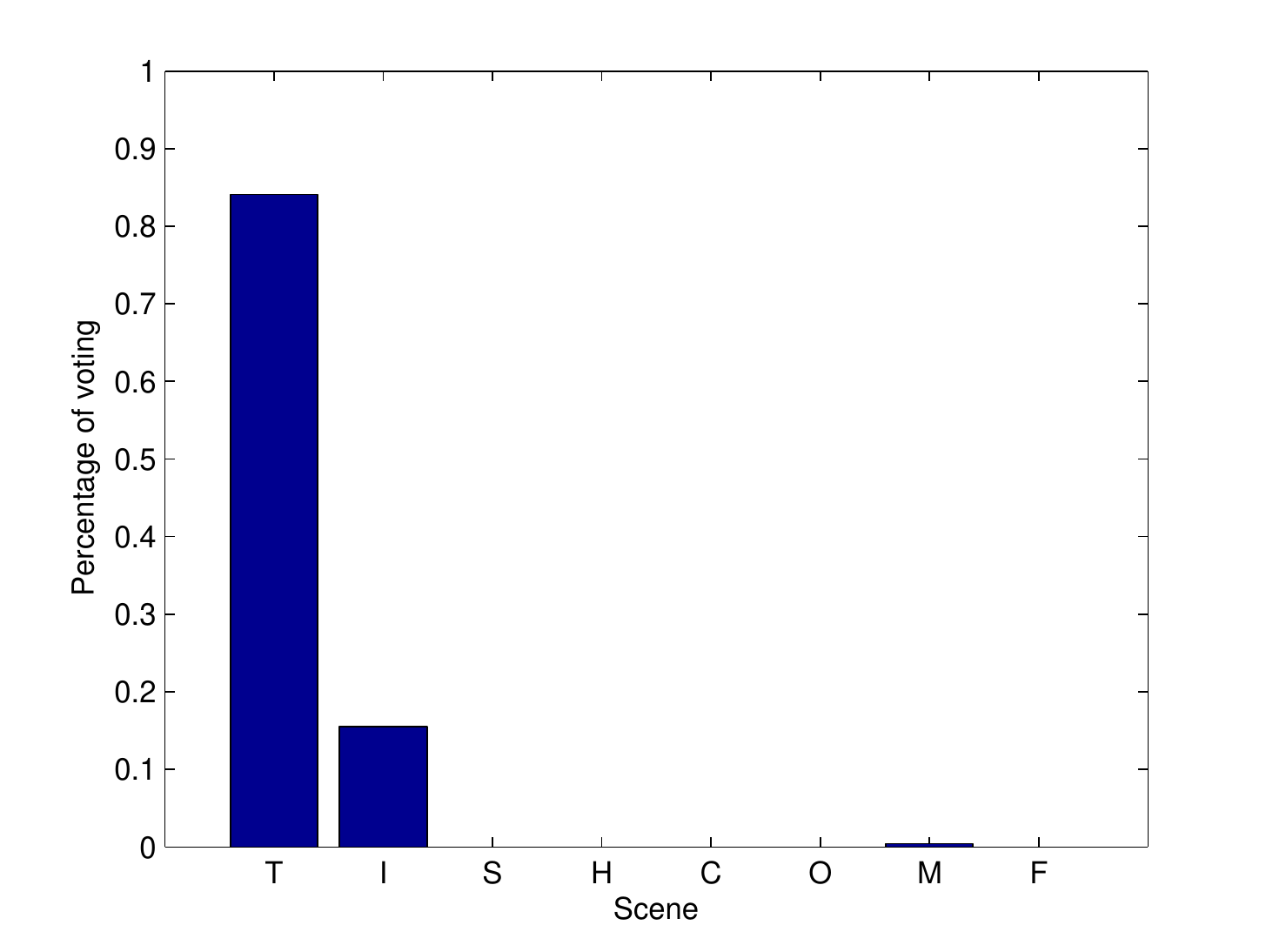}
\label{pic_obsSurvey2}}
\subfigure[Result of FQRC, $W_d$]{\includegraphics[scale=0.35]{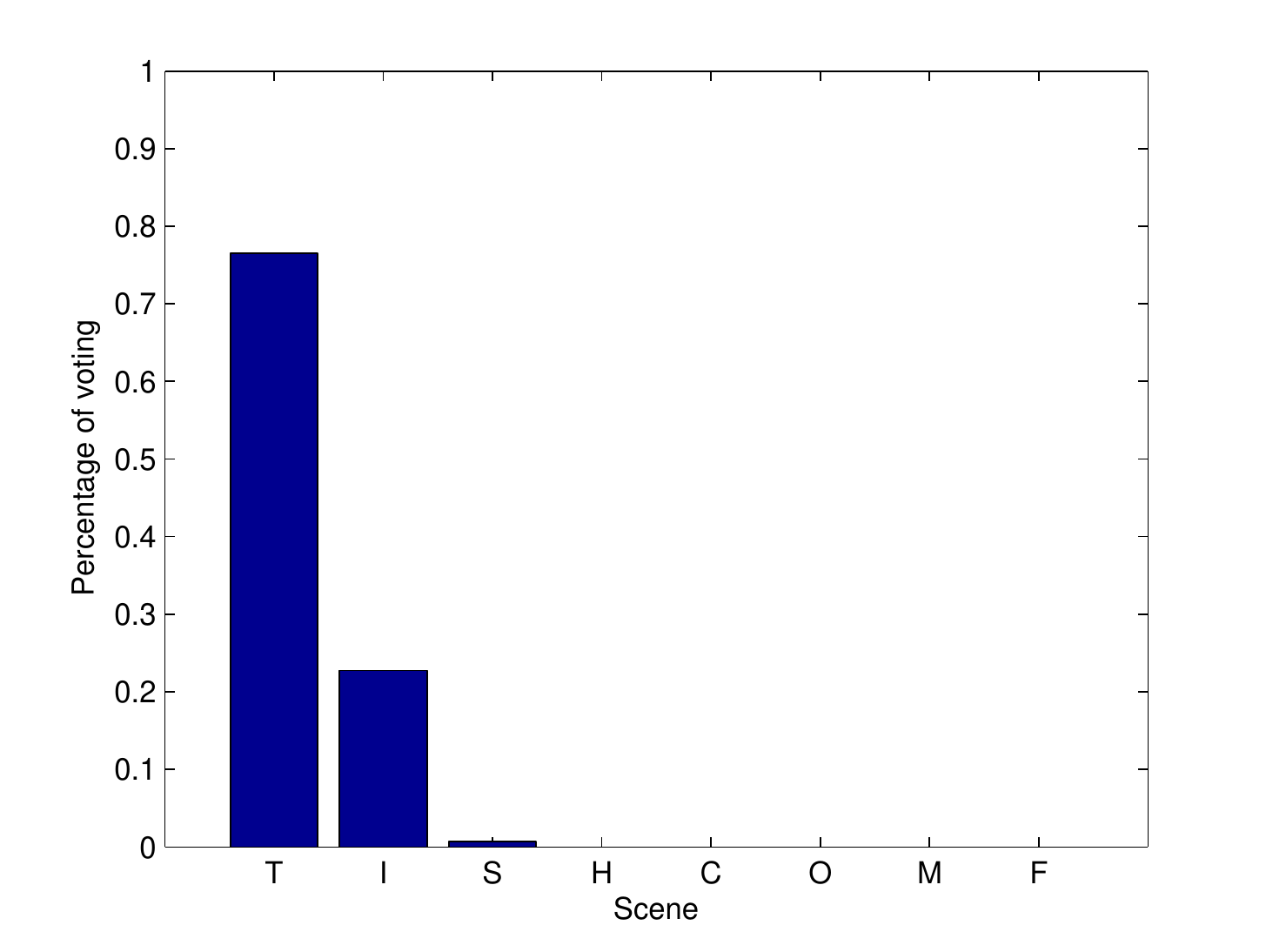}
\label{pic_obsComp2}}
\subfigure[Scene image 3]{\includegraphics[scale=0.5]{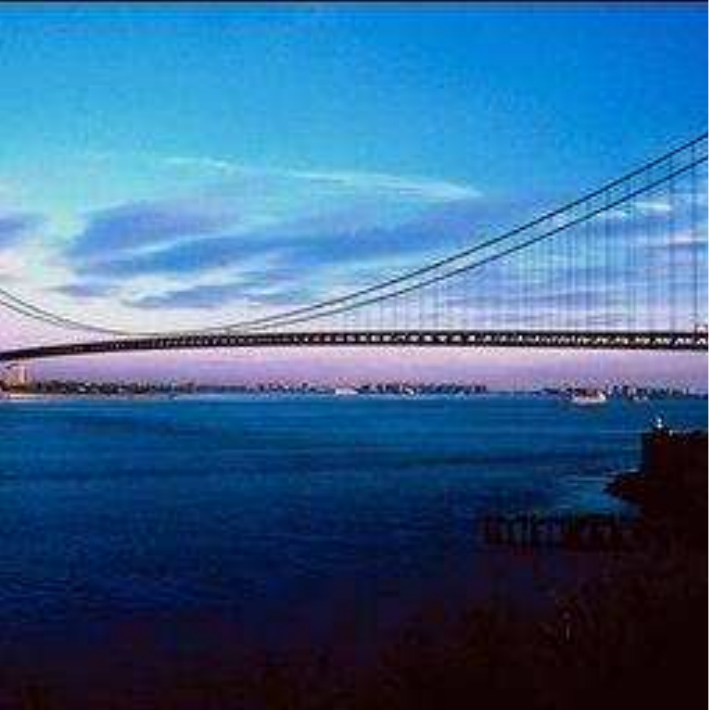}
\label{pic_obs4}}
\subfigure[Result of online survey, $Y_d$]{\includegraphics[scale=0.35]{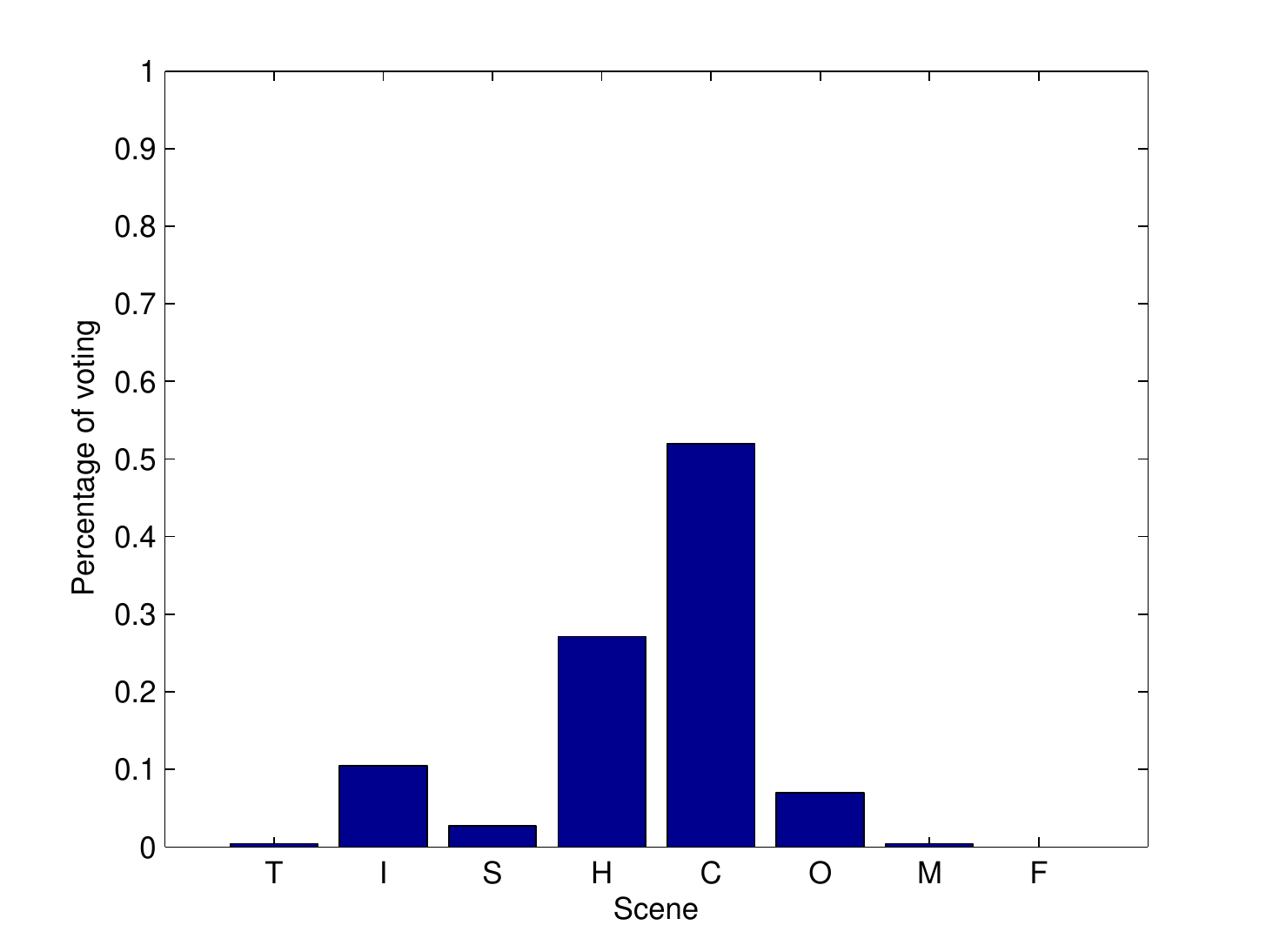}
\label{pic_obsSurvey4}}
\subfigure[Result of FQRC, $W_d$]{\includegraphics[scale=0.35]{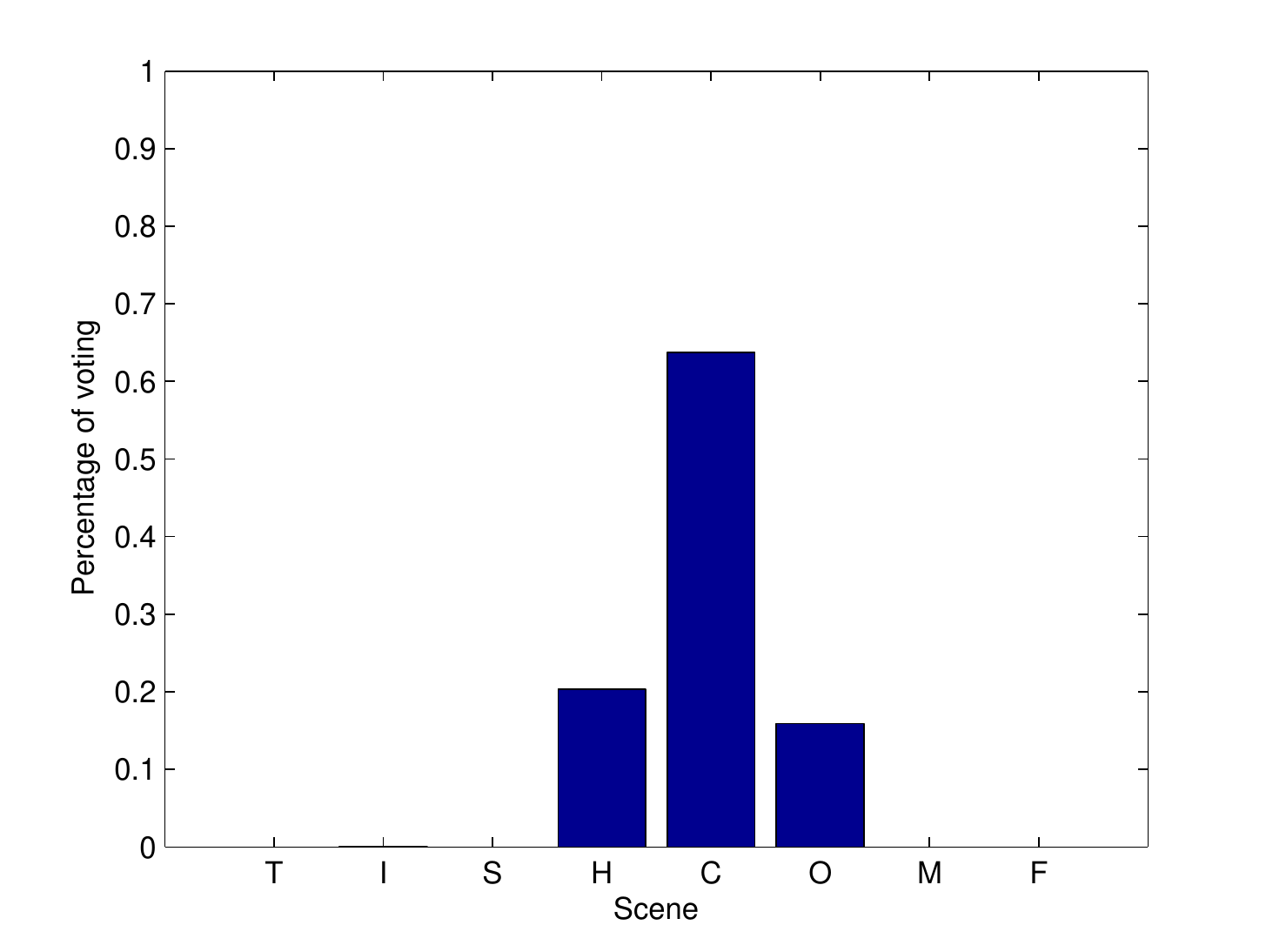}
\label{pic_obsComp4}}
\caption{Examples of the comparison between the results of online survey and FQRC (`Tallbuilding, T', `Insidecity, I', `Street, S', `Highway, H', `Coast, C', `Opencounty, O', `Mountain, M' and `Forest, F'). These results had shown that our proposed approach is very close to the human reasoning in scene understanding.}
\label{fig:obs}
\end{figure*}

We show the corresponding results from the online survey and our proposed FQRC in Fig. \ref{fig:obs}. Here, we can clearly see that the outcomes from both solutions are almost similar in terms of the ranking and the voting distributions. For instance, in Fig \ref{pic_obs2}, majority choose ''Tallbuilding'' (84.2\%) and follow by ''Insidecity'' (15.4\%). This is nearly close to the reading computed from FQRC where ''Tallbuilding'' is 76\% and ''Insidecity'' hold 22.7\%. 

However, one should understand that, this is almost impossible to obtain exactly the same values to the survey result due to the subjective nature of human decision making. What surprised us from the observation is the ranking of the distribution are very close to the results from FQRC compared to the survey. For example, in
Fig. \ref{pic_obs2}, by considering only the 'hit' labels for both results (`Tallbuilding, C' and `Insidecity, I'), the order of the distribution for FQRC computed result is $T>I$ which is similar to the survey results, where $T>I$ although the values are not exactly the same.  

From this observation, we can draw a preliminary conclusion that our proposed approach can emulate human reasoning in classifying scene images. To further validate this, quantitative evaluation is done in the following context. \\

\subsubsection{Quantitative Evaluation}
\label{QE}
In order to show that our proposed method is able to model the inconsistency of human decision making, we perform a quantitative evaluation using several evaluation criteria as explained in following contexts.
\\

\textbf{$\alpha$-Evaluation}. 
Evaluation of multi-label classification results is more complicated compared to binary classification because a result can be fully correct, partly correct, or fully incorrect. By using the example given by \cite{Boutell_Luo_Shen_Brown_2004}, let's assume we have classes $c_1,c_2,c_3$ and $c_4$. Take an example belongs to classes $c_1$ and $c_2$, we may get one of the results below:

\begin{itemize}
  \item $c_1$, $c_2$ (fully correct),
  \item $c_1$ (partly correct),
  \item $c_1$, $c_3$ (partly correct),
  \item $c_1$, $c_3$, $c_4$ (partly correct),
  \item $c_3$, $c_4$, (fully incorrect)
\end{itemize}

Herein, we wish to measure the degree of correctness of those possible results with their proposed $\alpha$-Evaluation. The score is predicted by the following formula:

\begin{equation}
score(W^b_d)=\left( 1-\frac{|\beta M_d+\gamma Q_d|}{|Y^b_d\cup W^b_d|} \right)^\alpha 
\label{eq:alphaEva}
\end{equation}

\noindent where $Y^b_d$ is the set of ground truth labels for the image sample $d$ in binary form ($Y_d > 0$) and $W^b_d$ is the set of prediction labels from the FQRC in binary form ($W_d > 0$). Also, $M_d=Y^b_d-W^b_d$ (missed labels) and $Q_d=W^b_d-Y^b_d$ (false positive labels). $\alpha,\beta$ and $\gamma$ are constraint parameters as explained in \cite{Boutell_Luo_Shen_Brown_2004}. In our evaluation, we select $\alpha=0.5, \beta=1$ and $\gamma=1$ and we calculate the accuracy rate of $D$.

\begin{equation}
accuracy_D=\frac{1}{|D|}\sum_{d\in D} score(W^b_d)
\label{eq:accD}
\end{equation}

\noindent where higher accuracy reflects better reliability of the FQRC because the `hit' label is almost similar to the survey results. 
\\

\textbf{Cosine similarity measure}.
Here, we would like to investigate the similarity of the histogram obtained from the survey and the FQRC, respectively by matching the pattern of the distributions. Cosine similarity measure has been employed for this purpose. First, we calculate the cosine distance (\ref{eq:cosine}) of the histogram distributions of each scenery image. Then we compute the average value of the similarity (\ref{eq:similarity}) to get the overall performance.

\begin{equation}
distance(W_d) = cos\Theta = \frac{Y_d\cdot W_d}{\left \| Y_d \right \|\left \| W_d \right \|}
\label{eq:cosine}
\end{equation}

\noindent The average similarity value for D;

\begin{equation}
similarity_D = \frac{1}{|D|}\sum_{d\in D} (1-distance(W_d))
\label{eq:similarity}
\end{equation}

\noindent where larger value of $similarity_D$ indicates higher similarity.
\\

\begin{figure}[htbp]
\centering
\includegraphics[scale=0.45]{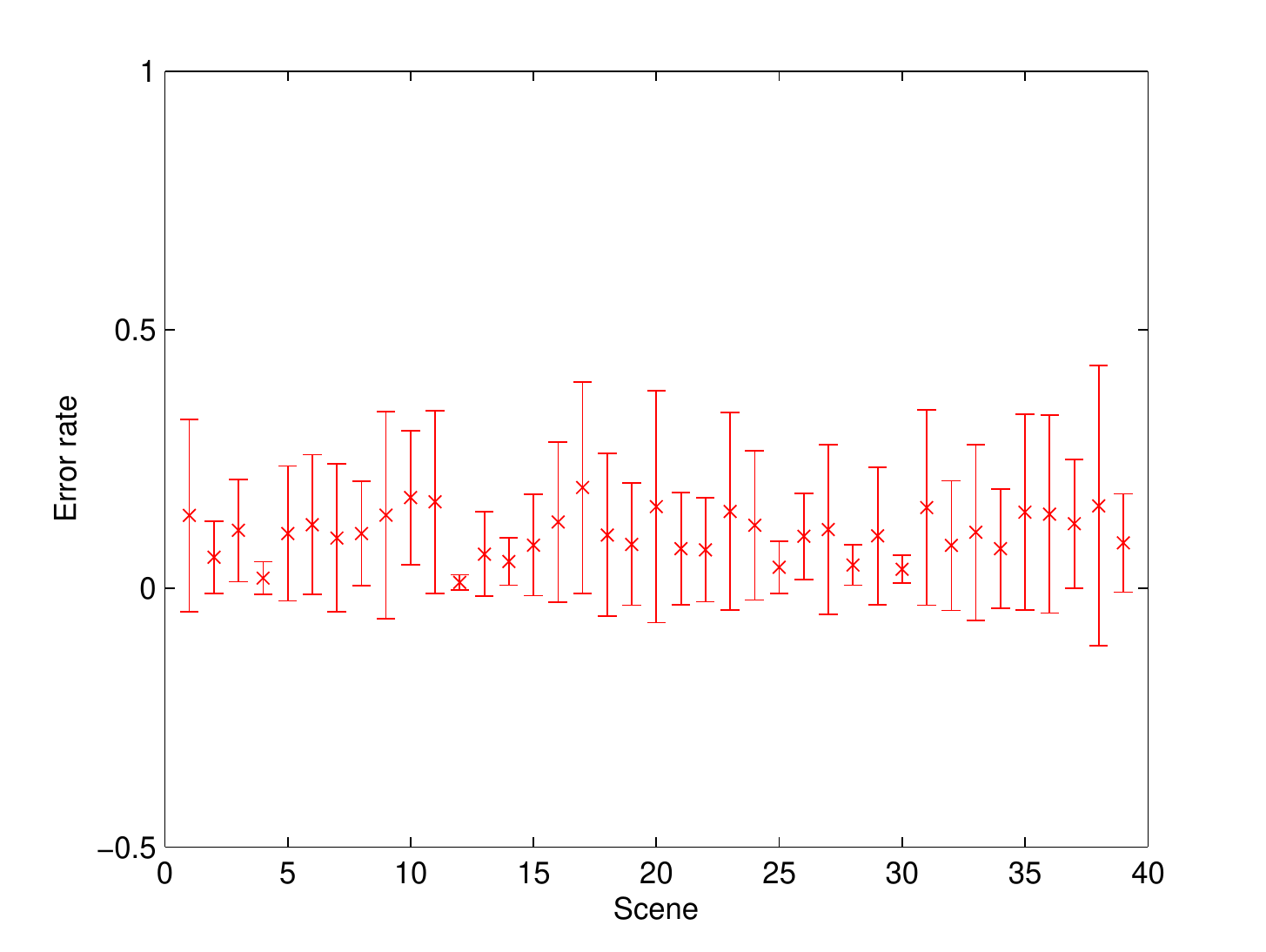}
\caption{Error bar of FQRC results compared to the online survey results for each scene image.}
\label{fig:err}
\end{figure}

\textbf{Error rate calculation}.
In this section, we investigate how much the computed result from the FQRC is deviated from the survey results. In order to achieve this, we obtain the error vector by subtracting both of the histogram distributions  (\ref{eq:errVec}). Then we calculate the mean and standard deviation of the error vector to observe the range of error as shown in Fig. \ref{fig:err}.

\begin{equation}
err(W_d) = |W_d-Y_d|
\label{eq:errVec}
\end{equation}

For the overall judgment in error rate, we compute the average standard deviation of the error values obtained from the scene images. Smaller value indicates less deviation of our results from the survey results.

All the three evaluation criteria are tested by a comparison between
the survey results (with and without $\alpha$-cut) and the proposed FQRC. The results are shown in Table \ref{table:eva1}.

\begin{table}[htbp]
	\centering
		\caption{Quantitative Evaluation of FQRC compared to online survey results}
		\label{table:eva1}
{\renewcommand{\arraystretch}{1.5}
\resizebox{8cm}{!} {
		\begin{tabular}{|c||c|c|c|c|}
			\hline
  		Scene & $\alpha$-evaluation & similarity & error \\ 
  		& (accuracy) & & (Average) \\ \hline \hline
			Without $\alpha$-cut & 0.75 					& 0.72 			& 0.13 				\\ \hline
			With $\alpha$-cut (1$\%$) & 0.79 					& 0.72 			& 0.13 				\\ \hline
		
		\end{tabular}}
		}
\end{table}

From the results in Table \ref{table:eva1}, we could observe reasonable output from these three evaluation criteria. The accuracy is above $70\%$, which indicates that the computational results using the FQRC is almost mimicking human reasoning in decision making where the `hit' label is highly matched with the answer from the survey. The high similarity here shows that our approach is able to provide an outcome that is similar to a human decision in terms of voting distribution and ranking. 

Based on the qualitative and quantitative results, we clarify that scene images are non-mutually exclusive and the state-of-the-art approach that uses binary classifier to deduce an unknown image to a specific class is impractical. Besides that, our proposed FQRC has proven its effectiveness as a remedy for this situation based on the comparison with the online survey results.

\begin{figure*}[htbp]
\centering
\subfigure[]{\includegraphics[height=0.18\linewidth, width=0.12\linewidth]{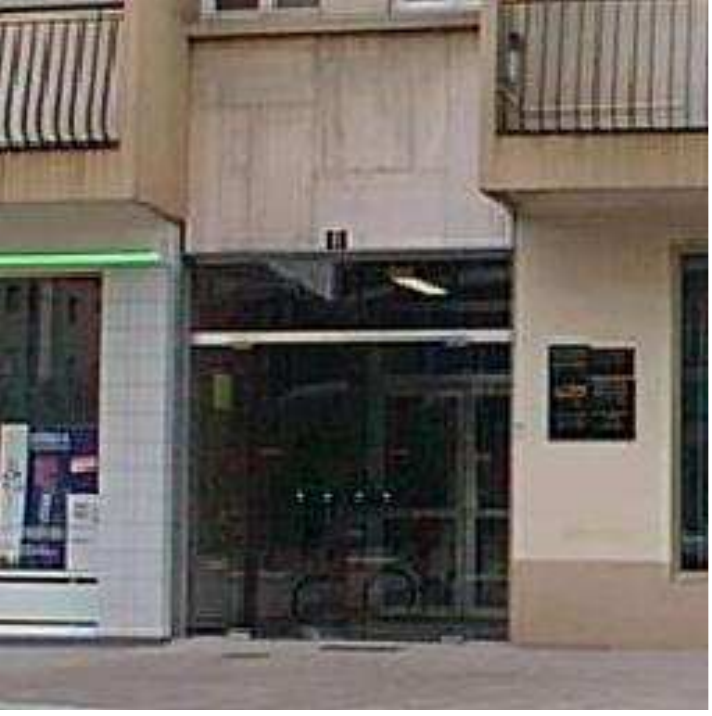}
\label{image4class_a}}
\subfigure[]{\includegraphics[height=0.18\linewidth, width=0.12\linewidth]{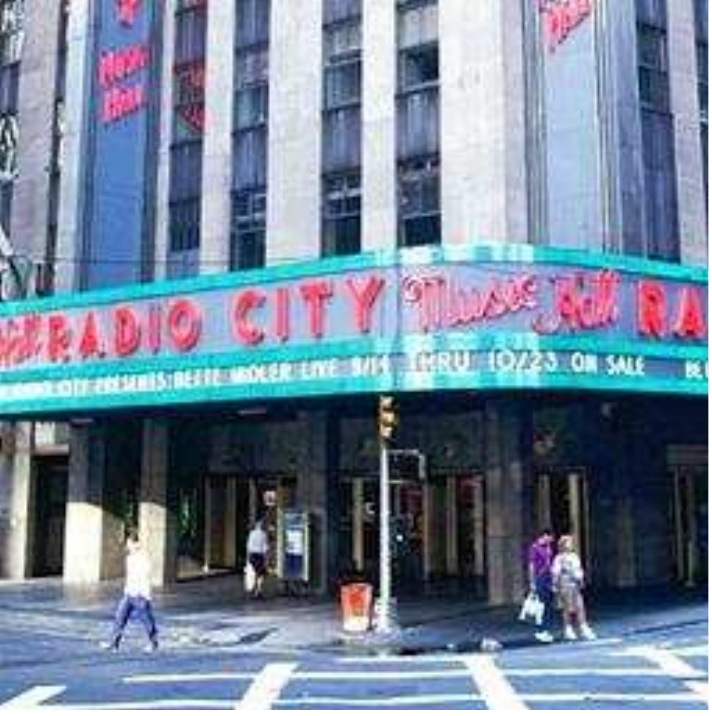}
\label{image4class_b}}
\subfigure[]{\includegraphics[height=0.18\linewidth, width=0.12\linewidth]{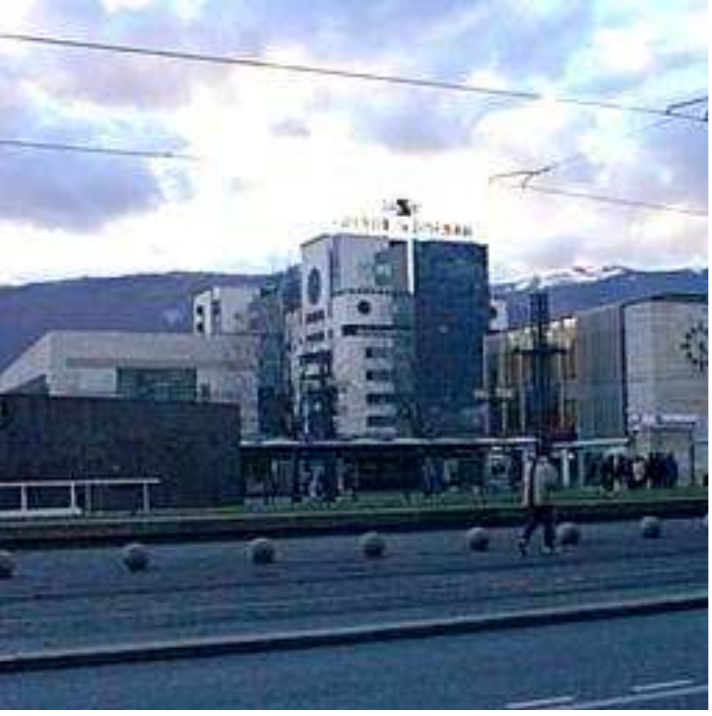}
\label{image4class_c}}
\subfigure[]{\includegraphics[height=0.18\linewidth, width=0.12\linewidth]{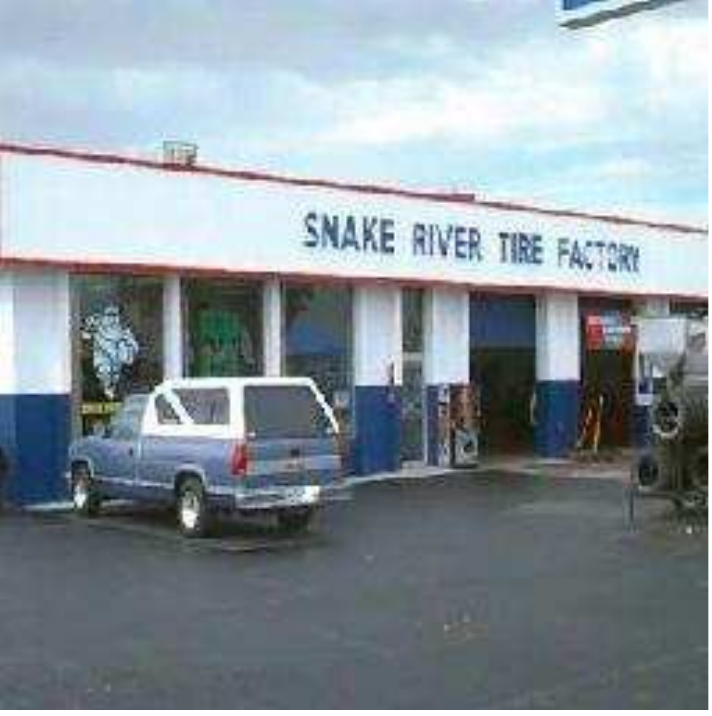}
\label{image4class_d}}
\subfigure[]{\includegraphics[height=0.18\linewidth, width=0.12\linewidth]{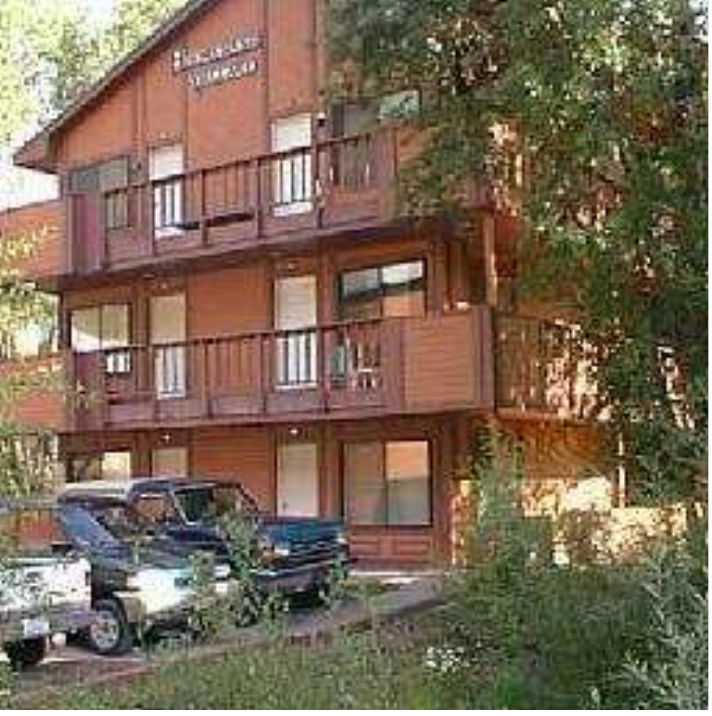}
\label{image4class_e}}
\subfigure[]{\includegraphics[height=0.18\linewidth, width=0.12\linewidth]{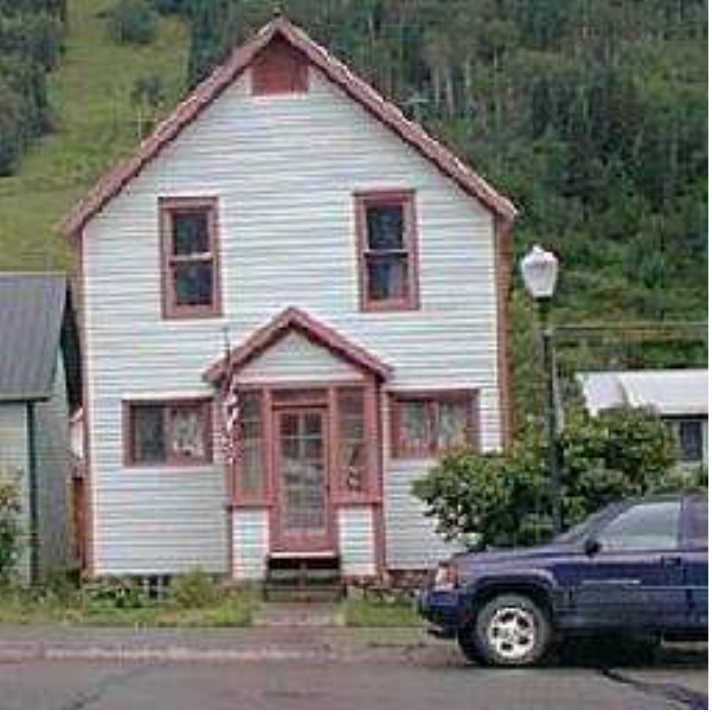}
\label{image4class_f}}
\subfigure[]{\includegraphics[height=0.18\linewidth, width=0.12\linewidth]{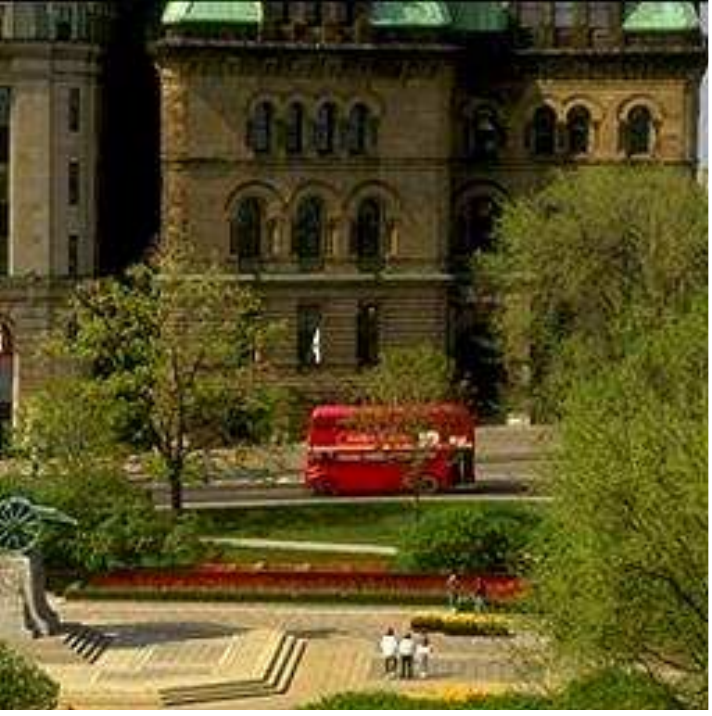}
\label{image4class_g}}
\caption{Example of images of Insidecity.}
\label{fig:image4class}
\end{figure*}

\subsection{Feasibility of FQRC}
\label{Exp2}

\begin{table}[htbp]
	\centering
		\caption{Comparison of the FQRC with the other
                  classifiers in terms of scene understanding}
		\label{Classifier}
{\renewcommand{\arraystretch}{1.3}
		\begin{tabular}{|c|c|c|c|c|}
			\hline
  		 	Classifier & Multi-label & Multi-class & Multi-dimension & Ranking		\\ \hline \hline
			KNN & - & \checkmark & -	& -	\\ \hline
			SVM & - & - & \checkmark	& -		\\ \hline
			\cite{noitacol} & - & \checkmark & \checkmark & -\\ \hline
			\cite{Boutell_Luo_Shen_Brown_2004} & \checkmark & \checkmark & \checkmark & - \\ \hline
			\textbf{FQRC} & \checkmark & \checkmark & \checkmark & \checkmark\\ \hline
		\end{tabular}}
\end{table}

In this experiment, we test the feasibility of our proposed FQRC in terms of multi-label, multi-class, multi-dimension and ranking. The explanation for each of the abilities is as below:

\begin{itemize}
  \item \textbf{Multi-label} - the classification outputs are associated with a set of labels
  \item \textbf{Multi-class} - the classifier that supports more than two classes, $K>2$ in a single classification task
  \item \textbf{Multi-dimension} - the classifier that supports more than two features, $J>2$ in a single classification task
  \item \textbf{Ranking} - higher interpretation of the classification results by reordering the inference outcome.
\end{itemize}

Table \ref{Classifier} shows how the FQRC distinguishes itself from the other classifiers and each of the capabilities has been clarified with the succeeding experiments in following sections. \\

\subsubsection{FQRC with 2 attributes and 4 scene classes (Multi-label \& Multi-class)}

From the comparison results show in Table
\ref{tabel:4class2attr_output}, it can be observed that one drawback
of \cite{chernhong} is it provides similar results on certain images,
which is very absurd as all the corresponding images are so different
from each other and imply that each of the images has its own value of
attributes, which should be different from other images. Our proposed
approach, in contrast, is able to model this behavior and provides an
output that is closer to human perspective. Apart from that, the
confident values inferred from our approach are more reasonable
compared to \cite{chernhong}, for example, in
Fig. \ref{image4class_e}, even for subjective judgment, we will
consider that the confident level of this image belonged to
`Insidecity' is higher than the 'Forest'. This improvement is mainly
from the proposed 4-tuple fuzzy membership learning algorithm.

\begin{table*}[htbp]
	\centering
		\caption{Inference output with 2 attributes and 4 classes of Fig. \ref{fig:image4class}}
		\label{tabel:4class2attr_output}
{\renewcommand{\arraystretch}{1.5}
\resizebox{14cm}{!} {
		\begin{tabular}{|c||c|c|c|c||c|c|c|c|}
			\hline
  		\multirow{2}{*}{Scene} & \multicolumn{4}{|c||}{\textbf{FQRC}} & \multicolumn{4}{|c|}{\cite{chernhong}} \\ \cline{2-9}
  		 & Insidecity & Coast & Opencountry & Forest & Insidecity & Coast & Opencountry & Forest \\ \hline \hline
			\ref{image4class_a} & 0.9280 & 0 & 0 & 0.0720 & 1 & 0  & 0 & 0 \\ \hline
			\ref{image4class_b} & 1 & 0 & 0 & 0 & 1 & 0 & 0 & 0 \\ \hline
			\ref{image4class_c} & 0.5068 & 0.1587 & 0.3344 & 0 & 0.7273 & 0.2727 & 0 & 0 \\ \hline
			\ref{image4class_d} & 0.6845 & 0 & 0.3155 & 0 & 0.7273 & 0.2727 & 0 & 0 \\ \hline
			\ref{image4class_e} & 0.5296 & 0.0483 & 0 & 0.4221 & 0.1250 & 0 & 0.1250 & 0.7500 \\ \hline
			\ref{image4class_f} & 0.5872 & 0.0146  & 0.007 & 0.3911 & 0.8235 & 0 & 0 & 0.1765 \\ \hline
			\ref{image4class_g} & 0.5561 & 0.0264 & 0 & 0.4175 & 0.8235 & 0 & 0 & 0.1765 \\ \hline			
		\end{tabular}}
		}
\end{table*}

\subsubsection{FQRC with 6 attributes and 4 scene classes (Multi-dimension)}

In this testing, our proposed framework shows the strength of
performing multi-dimensional classification compare to
\cite{chernhong} where we employ 6 attributes instead of 2 to perform
the classification tasks. The 6 attributes are the score values of
`Nature', `Open', `Perspective', `Size-Large', `Diagonal-Plane', and
`Depth-Close', respectively. Using the similar testing images as in
Fig. \ref{fig:image4class}, the classification results from the FQRC are shown in Table \ref{tabel:4class6attr_output}.

\begin{table}[htbp]
	\centering
		\caption{Inference output with 6 attributes and 4 classes of Fig. \ref{fig:image4class}}
		\label{tabel:4class6attr_output}
{\renewcommand{\arraystretch}{1.4}
\resizebox{8cm}{!} {
		\begin{tabular}{|c||c|c|c|c|}
			\hline
  		Scene & Insidecity & Coast & Opencountry & Forest \\ \hline \hline
			\ref{image4class_a} & 1 					& 0 			& 0 					& 0 			\\ \hline
			\ref{image4class_b} & 1 					& 0 			& 0 					& 0 			\\ \hline
			\ref{image4class_c} & 0.6722 			& 0.1001 	& 0.2277 			& 0 			\\ \hline
			\ref{image4class_d} & 0.9179 			&	0 			&	0.0821			&	0 			\\ \hline
			\ref{image4class_e} & 0.5188 			&	0 			&	0						&	0.4812 	\\ \hline
			\ref{image4class_f} & 0.8411			& 0				&	0.0013			& 0.1575	\\ \hline
			\ref{image4class_g} & 0.5936			&	0				&	0						&	0.4064	\\ \hline			
		\end{tabular}}
		}
\end{table}

As we compare the result between Table \ref{tabel:4class2attr_output}
and \ref{tabel:4class6attr_output}, one can observe that the result
using six attributes are more reasonable than two attributes,
especially in
Fig. \ref{image4class_a},\ref{image4class_e},\ref{image4class_f}, and
\ref{image4class_g}, respectively. Here in the case of
Fig. \ref{image4class_e}, with the used of six attributes as to two
attributes, the result improved in term of eliminated the noise which
is the `Coast' class that should never been an option for this
particular image. However, we observe that the values of confident of
Fig. \ref{image4class_e} in `Insidecity' and `Forest' have change
significantly. But still, they are in the manner where the confident
level of `Insidecity' is more than `Forest' which is matched to the
subjective judgment. 

Slight changes of these results were incurred as a resultant from the
additional of the number of attributes into the classification
framework. In fact, more attributes tend to increase the uniqueness of
one class from another and this indirectly has increased the
discriminative strength of the classifier. However, it is almost
impossible to find the optimum attributes (or features) that are best
to distinguish one class from another classes especially in the scene
understanding task (non-mutually exclusive case). Furthermore, using
excessive attributes in the algorithm will increase the computational
cost. Therefore, our proposed framework considers a more generative
way that provides a good tradeoff between the multi-dimensional classification
capability and the performance of the classification task.

\subsubsection{FQRC in ranking (Ranking ability)}
The goal of this experiment is to show the effectiveness of the
proposed FQRC in higher interpretation such as ranking by classifying
the possibility of an unknown image into the eight learned scene
classes with the correct ordering. To provide more information, we output it with some symbolic representation explained in Section \ref{IR} rather than classify it as one of the eight learned scene classes. The reason is we do not assume the scene classes are mutually exclusive and we understand that scene classes are arbitrary and possibly sub-optimal. 

\begin{table*}[htbp]
	\centering
		\caption{Inference output with 6 attributes and 8 classes of Fig. \ref{fig:image4class}}
		\label{table:8class6attr_output}
{\renewcommand{\arraystretch}{1.5}
\resizebox{14cm}{!} {
		\begin{tabular}{|c||c|c|c|c|c|c|c|c|}
			\hline
  		 	Scene Image						& Tallbuilding & Insidecity & Street & Highway & Coast & Opencountry & Mountain & Forest \\ \hline \hline
			\ref{image4class_a} & 0.4562	& 0.4562	& 0.0876	& 0				& 0				& 0				& 0				& 0				\\ \hline
			\ref{image4class_b} & 0.7644	& 0.2356	& 0				& 0				& 0				& 0				& 0				& 0				\\ \hline
			\ref{image4class_c} & 0				&	0.3339	&	0.0308	& 0.4725	&	0.0497	&	0.1131	& 0				& 0				\\ \hline
			\ref{image4class_d} & 0				&	0.5880	&	0.0499	&	0.3094	&	0				& 0.0526	&	0				&	0				\\ \hline
			\ref{image4class_e} & 0.0726	&	0.2631	&	0.4202	&	0				&	0				&	0				&	0				&	0.2440	\\ \hline
			\ref{image4class_f} & 0.1412	&	0.3456	&	0.4361	&	0				& 0				&	0.0005	&	0.0119	&	0.0647	\\ \hline
			\ref{image4class_g} & 0.0811	&	0.2826	&	0.4183	&	0				& 0				&	0				&	0.0245	&	0.1935	\\ \hline			
		\end{tabular}}
		}
\end{table*}

Table \ref{table:8class6attr_output} shows the sub-sample results
using randomly selected scene images from the `Insidecity' class.  The
visual appearances of these images are illustrated in
Fig. \ref{fig:image4class}. Herein, we can notice that (1) The FQRC is able to correctly classify each image which has the possibility (confident value, $r_k$) in `Insidecity' class. This is true as the benchmarking for these sub-sample images is selected from the Class = `Insidecity'. Nonetheless, our approach also discovered that each of these images can have possibility belongs to other classes. For instance, our approach discovered that Fig. \ref {image4class_a} has the possibility as `Tallbuilding' and `Street' class. Table \ref{table:8class6attr_interpretation} shows the symbolic representation of the ranking based on the results in Table \ref{table:8class6attr_output} where $q$ refers to scene image 11(a) to 11(g) in particular interpretation.

\begin{table}[htbp]
	\centering
		\caption{Interpretation of ranking of 6 attributes and 8 classes. Images are from the following categories: Tallbuilding(T), Insidecity(I), Street(S), Highway(H), Coast(C), Opencountry(O), Mountain(M), and Forest(F)}
		\label{table:8class6attr_interpretation}
{\renewcommand{\arraystretch}{1.5}
		\begin{tabular}{|c||c|}
			\hline
  		 	Scene Image							& Interpretation 				\\ \hline \hline
			\ref{image4class_a} & T$\equiv$I$>$S, q$\times$H,C,O,M,F				\\ \hline
			\ref{image4class_b} & T$\gg$I, q$\times$S,H,C,O,M,F							\\ \hline
			\ref{image4class_c} & H$>$I$>$O$>$C$>$S, q$\times$T,M,F					\\ \hline
			\ref{image4class_d} & I$>$H$>$O$>$S, q$\times$T,C,M,F						\\ \hline
			\ref{image4class_e} & S$>$I$>$F$>$T, q$\times$H,C,O,M						\\ \hline
			\ref{image4class_f} & S$>$I$>$T$>$F$>$M$>$O, q$\times$H,C				\\ \hline
			\ref{image4class_g} & S$>$I$>$F$>$T$>$M, q$\times$H,C,O					\\ \hline
		\end{tabular}}
\end{table}

\begin{figure*}[htbp]
\centering
 \subfigure[]{\includegraphics[height=0.2\linewidth, width=0.3\linewidth]{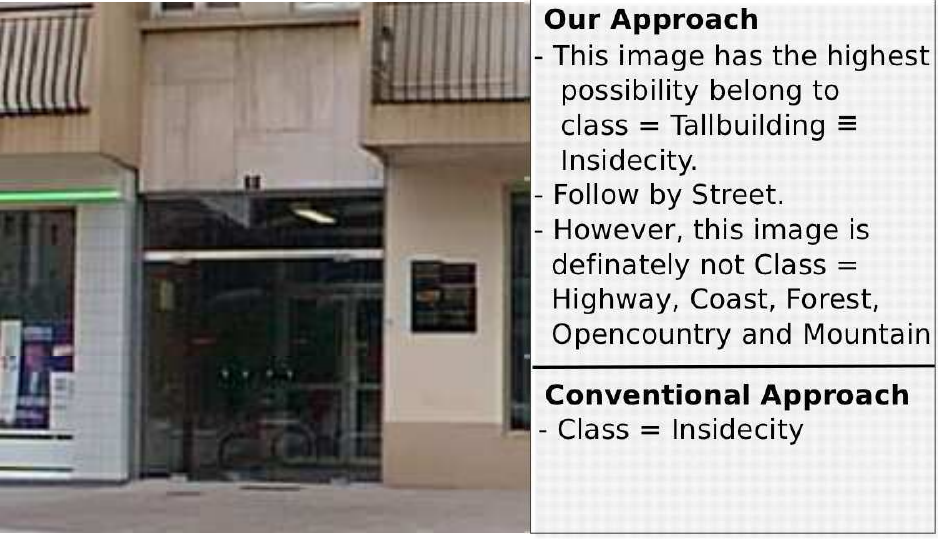}
 \label{image4class_a1}}
\subfigure[]{\includegraphics[height=0.2\linewidth, width=0.3\linewidth]{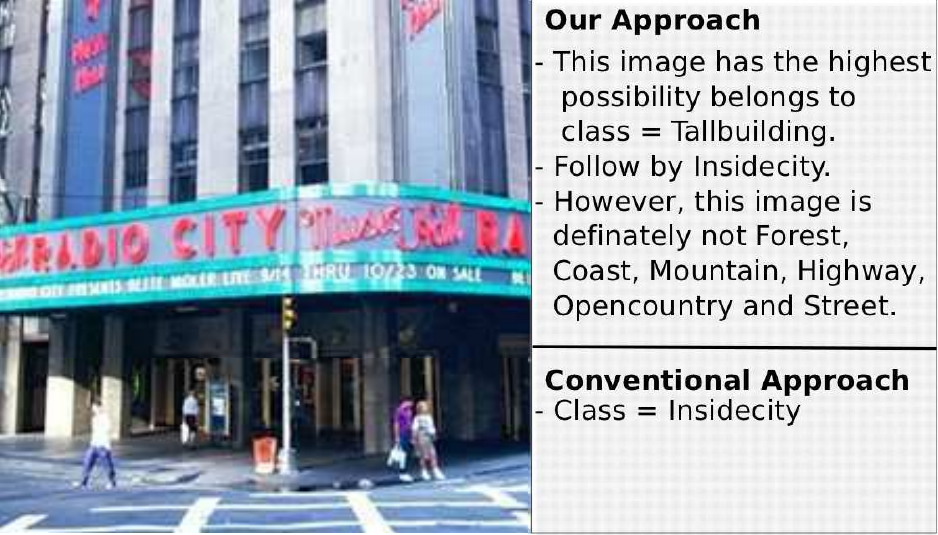}
\label{image4class_b1}}
\subfigure[]{\includegraphics[height=0.2\linewidth, width=0.3\linewidth]{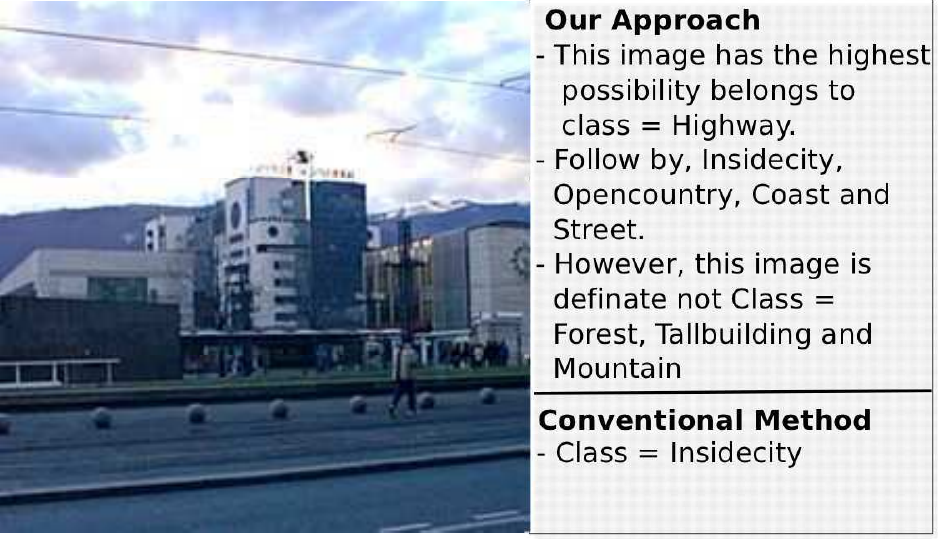}
\label{image4class_c1}}
 \subfigure[]{\includegraphics[height=0.2\linewidth, width=0.3\linewidth]{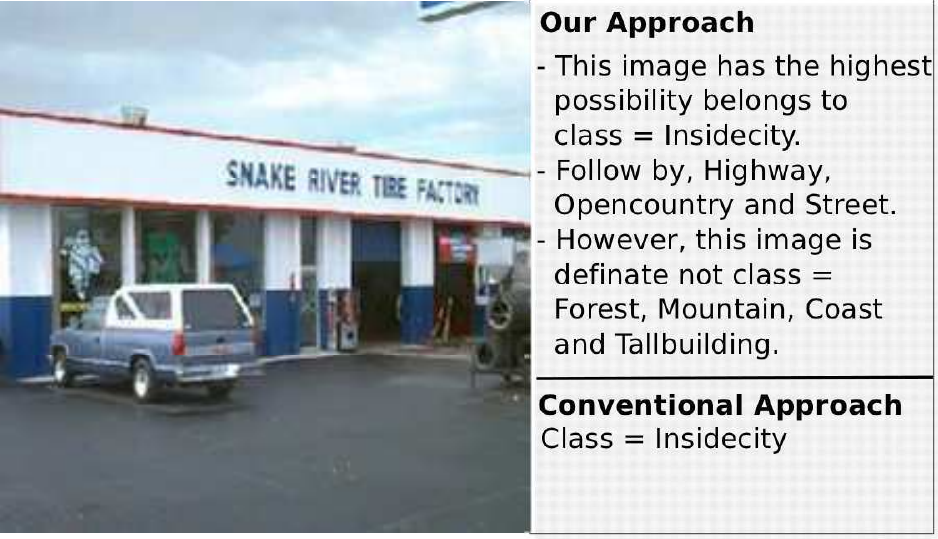}
 \label{image4class_d1}}
 \subfigure[]{\includegraphics[height=0.2\linewidth, width=0.3\linewidth]{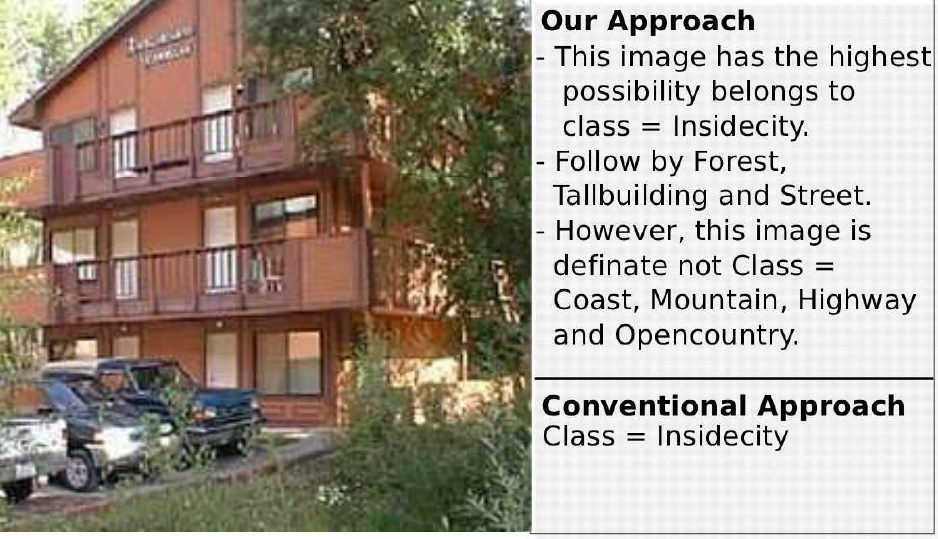}
 \label{image4class_e1}}
\subfigure[]{\includegraphics[height=0.2\linewidth, width=0.3\linewidth]{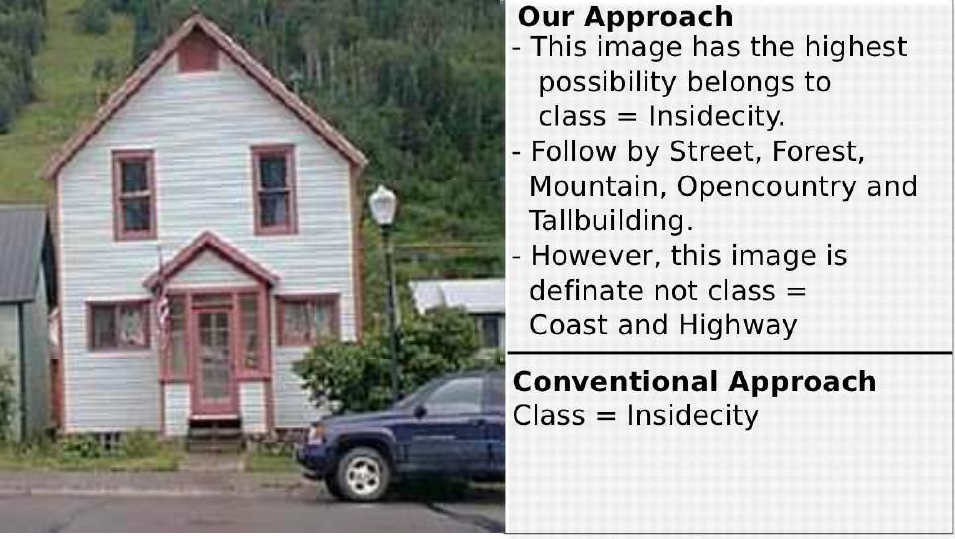}
\label{image4class_f1}}
\caption{Example of the ranking interpretation in textual descriptions.}
\label{fig:image4class1}
\end{figure*}

As mentioned in the context, our approach are capable of generate
purely linguistic descriptions where an image is described relative to
other categories. Fig. \ref{fig:image4class1} shows the examples. Echoing our quantitative results, we can qualitatively observe that the relative descriptions are more precise and informative than the binary ones.

\subsection{Comparison to state-of-the-art binary classifiers in single label classification task}
\label{Exp3}

One of the strengths of the FQRC is it provides the feasibility to
perform single-label classification task like other binary classifiers
as well as ranking as shown in the aforementioned subsection. To verify this,
here, we compare the FQRC against the state-of-the-art binary
classifiers such as K-nearest neighbor (KNN), Directed Acyclic Graph
SVM (DAGSVM) \cite{noitacol}, and Fuzzy least squares SVM (LSSVM)
\cite{Tsujinishi2003785}. In the FQRC, we have employed max aggregation, $z=max(r)$  to obtain the maximum confident value as binary classification results.

For a fair comparison, we perform the classification task with 2 attributes and 4 classes for all classifiers as KNN could not handle the multi-dimensional classification. In the configuration of each classifier in the comparison, we use conventional KNN with empirical chosen parameter $K=5$. As for DAGSVM \cite{noitacol} and LSSVM \cite{Tsujinishi2003785}, DAGSVM runs with RBF as kernel and margin parameter, $C=100$ using SMO training while LSSVM is implemented based on linear SVM with $C=2000$ and incorporates with the least square solution. 

\begin{figure*}[htbp]
\centering
\subfigure[Insidecity]{\includegraphics[height=0.3\linewidth, width=0.23\linewidth]{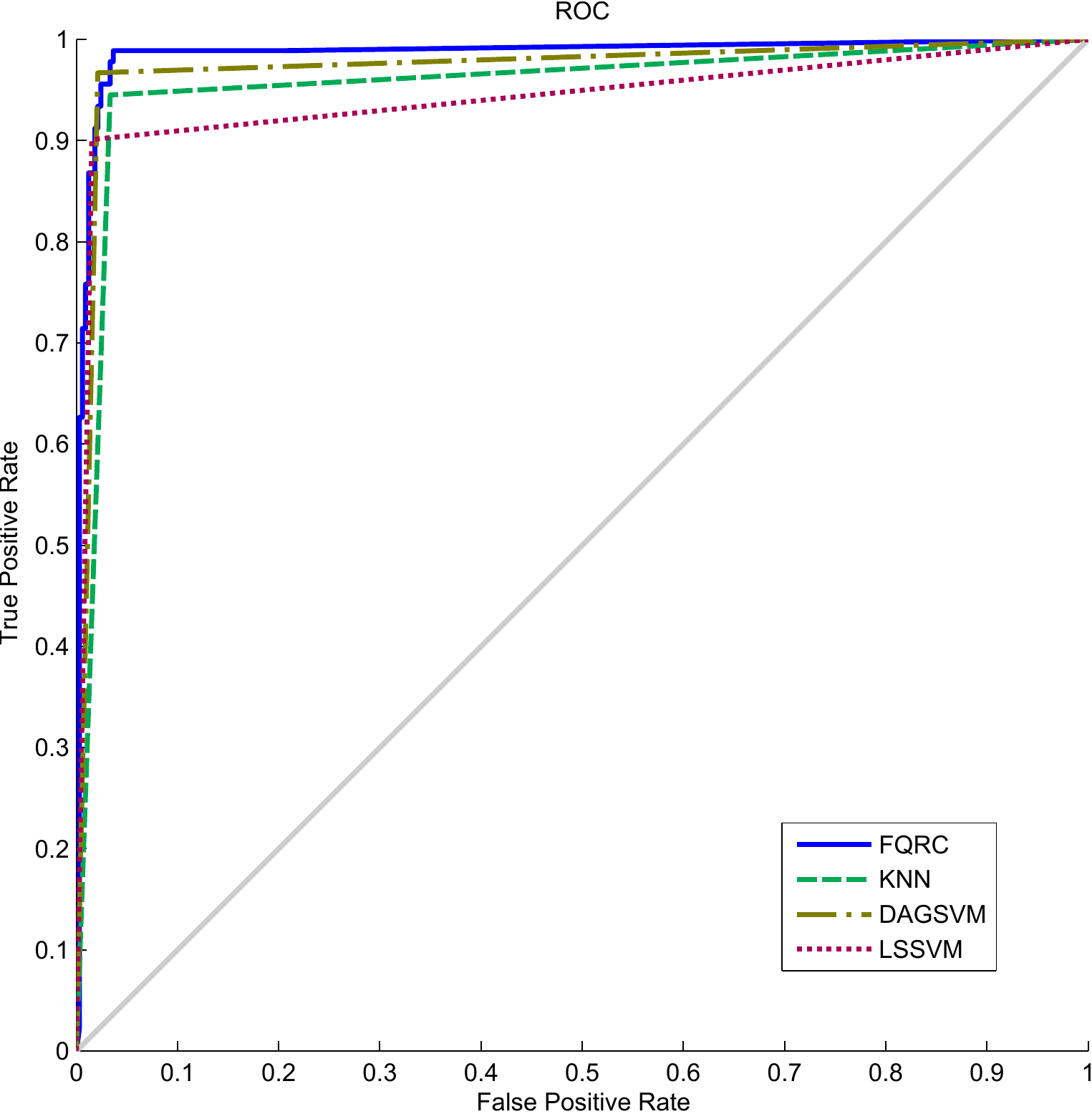}
\label{pic_ROC_4class2Attr_insidecity}}
\subfigure[Coast]{\includegraphics[height=0.3\linewidth, width=0.23\linewidth]{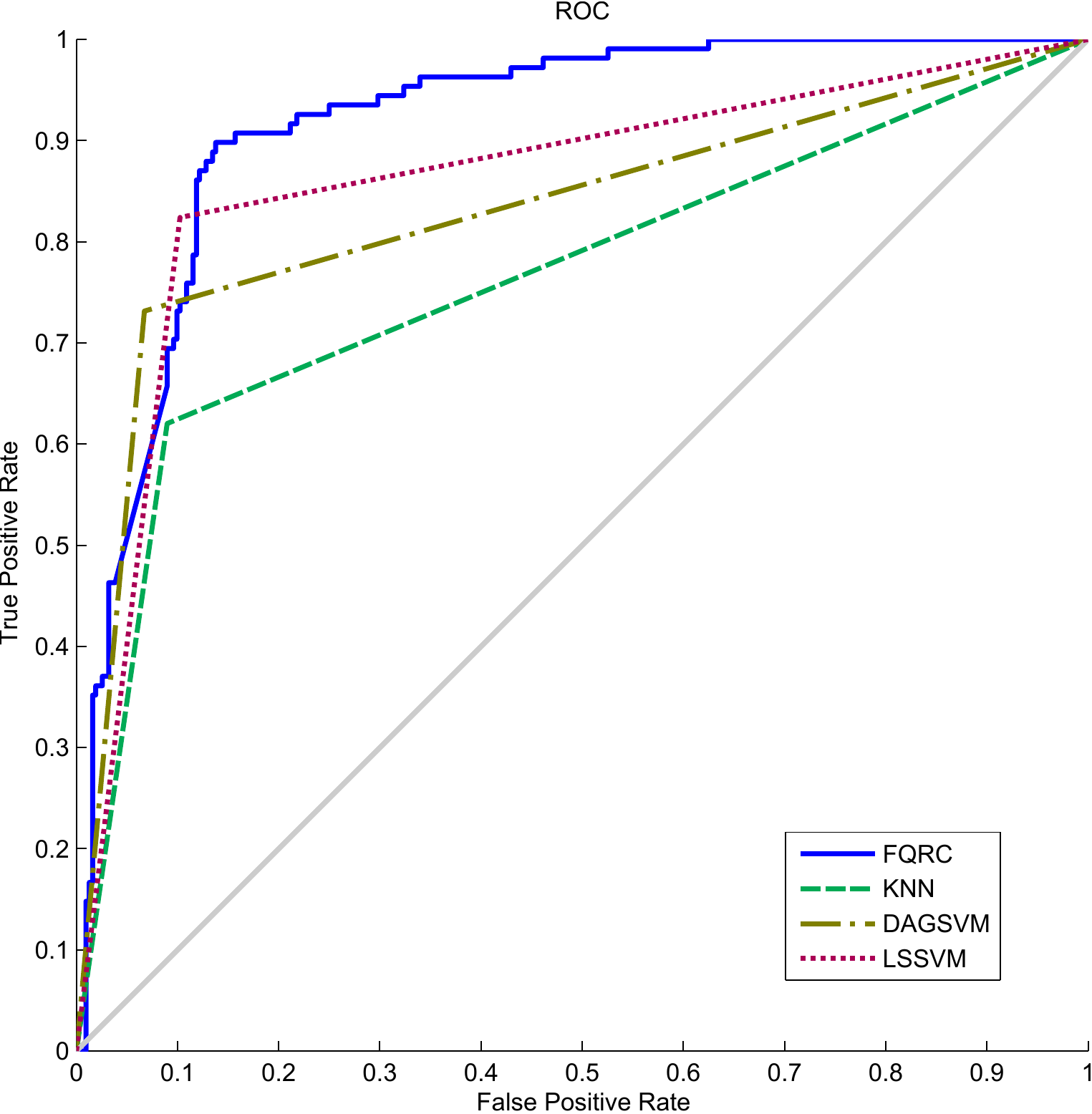}
\label{pic_ROC_4class2Attr_coast}}
\subfigure[Opencountry]{\includegraphics[height=0.3\linewidth, width=0.23\linewidth]{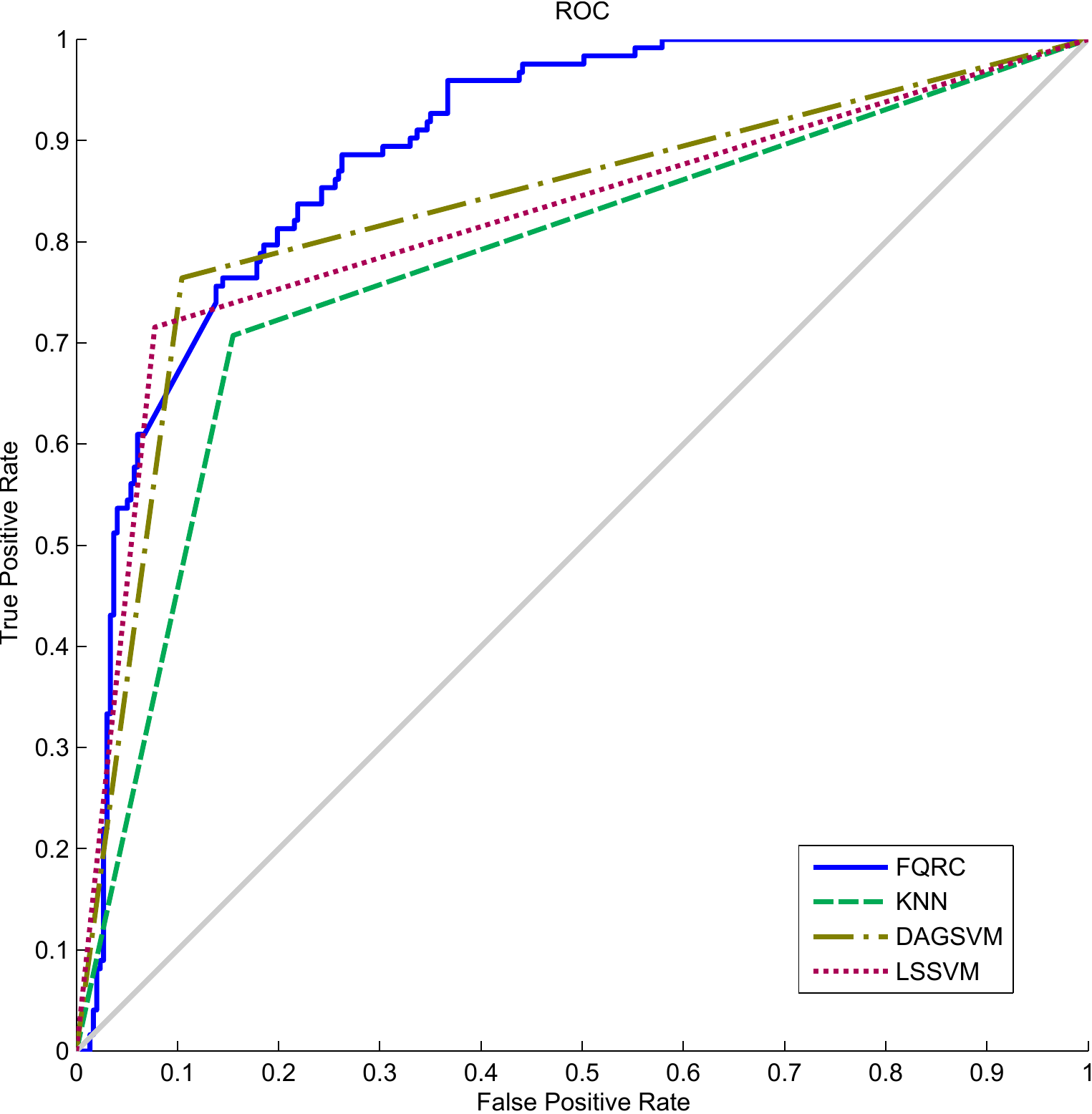}
\label{pic_ROC_4class2Attr_opencountry}}
\subfigure[Forest]{\includegraphics[height=0.3\linewidth, width=0.23\linewidth]{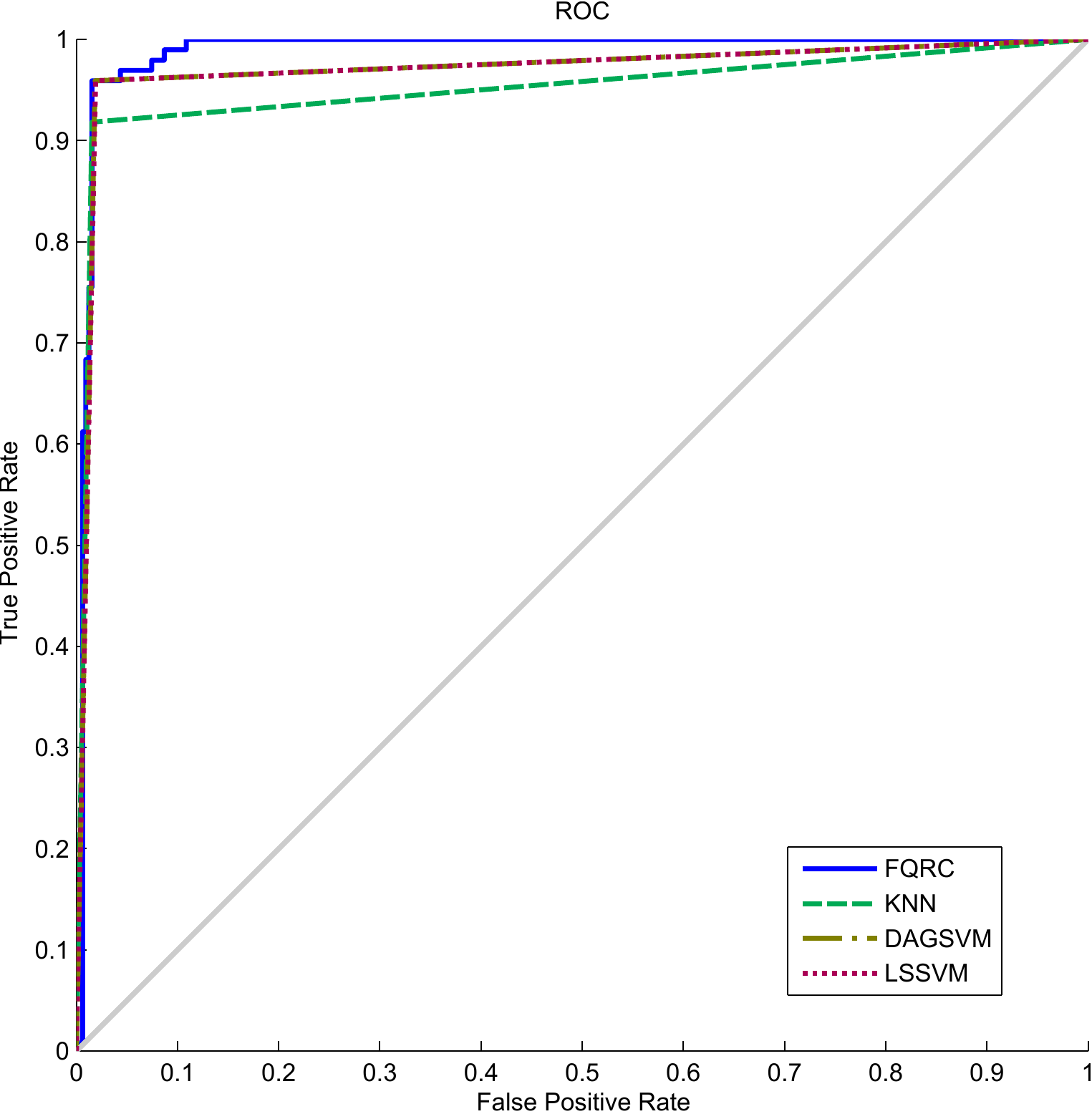}
\label{pic_ROC_4class2Attr_forest}}
\caption{ROC comparison between FQRC and the other binary classifiers.}
\label{fig:ROC}
\end{figure*}

\begin{figure}[htbp]
\centering
\includegraphics[scale=0.25]{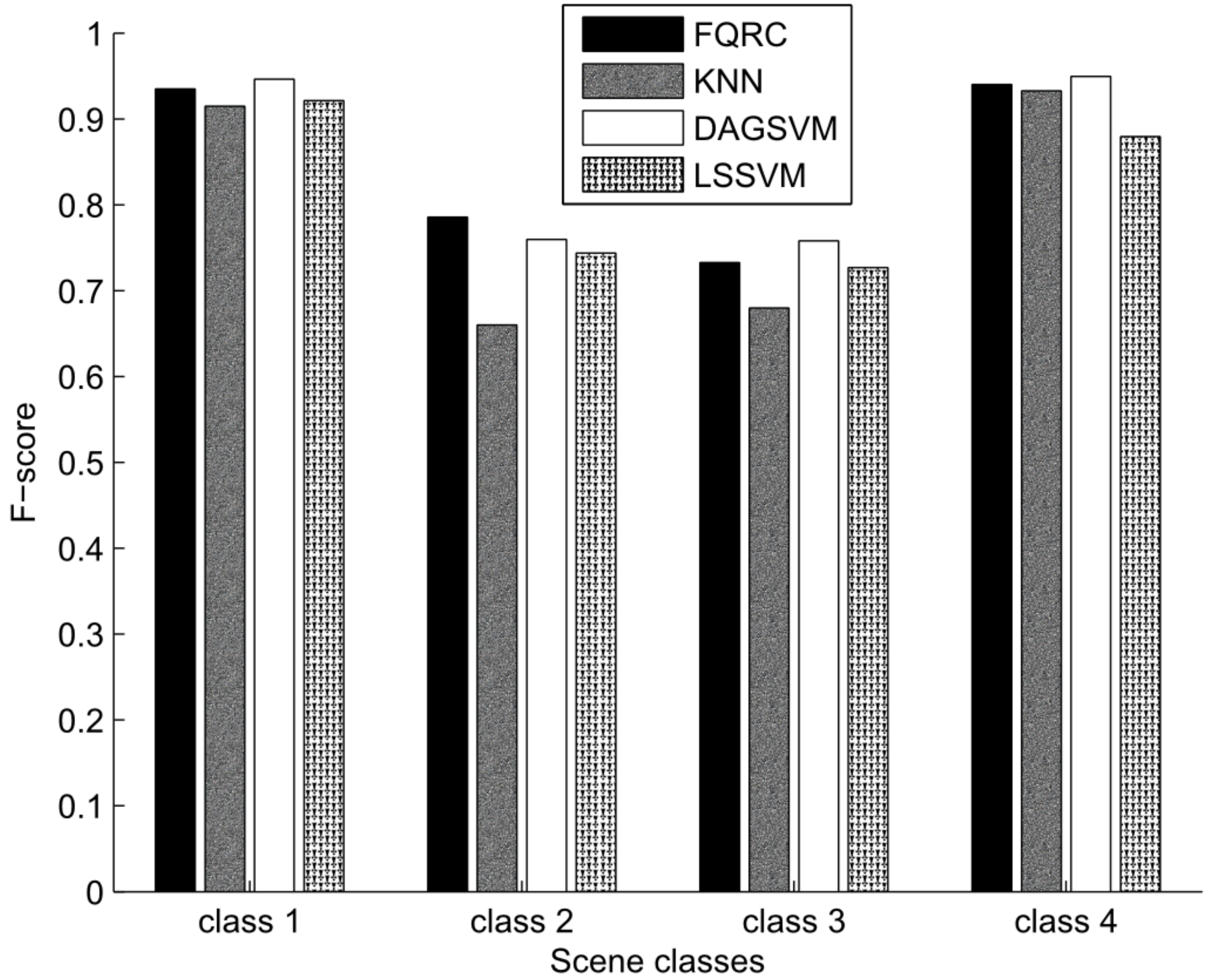}
\caption[F-score]{Comparison of F-score between the classifiers. Class 1 (Insidecity), Class 2 (Coast), Class 3 (Opencountry), and Class 4 (Forest).}
\label{fig:fs}
\end{figure}

The $F-score$ (Fig. \ref{fig:fs}) is calculated to show the accuracy of the classification task by comparing our FQRC and three other classifiers. In information retrieval literatures, the $F-score$ is often used for evaluating this quantity: 

\begin{equation}
F = \frac{{2\nu\rho}}{{\nu + \rho}}. 
\label{eq:Fscore}
\end{equation} 

The recall, $\rho$ and the precision, $\nu$ measure the configuration
errors between the ground truth and the classification result. For a good inference quality, both the recall and precision should have high values. The ROC graphs show in Fig. \ref{fig:ROC} is to evaluate the sensitivity of the classifiers while Fig. \ref{fig:fs} illustrates the F-score for each classification task. From both figures, we can notice that our proposed method is comparable with the KNN, DAGSVM, and LSSVM. Most of the time, FQRC outperforms other binary classifiers but is slightly inefficient as compared to DAGSVM.

One of the main reasons is DAGSVM used an efficient data structure to
express the decision node in the graph, and an improved decision
algorithm is used to find the class of each test sample and thus makes
the decision more accurate compared to other binary classifiers. In
short, DAGSVM is a discriminative classifier that was implemented and
trained to distinguish distinctly amongst the data where there is no
cross-over tolerance in the data distribution. This is in contrary to
the FQRC as a generative classifier to relief the ignorance of
non-mutually exclusive data. In conclusion, DAGSVM should be better in
most of the case compared to FQRC as a binary classifier.  However,
here, in this context, the objective is to show that one of the
strengths of FQRC is the capability to perform a single-label classification task while playing the role of ranking classifier, which yields comparable results with the other state-of-the-art binary classifiers.

\subsection{Comparison to state-of-the-art multi-label scene classification approaches}
\label{Exp4}

In order to show the effectiveness and efficiency of our proposed method, in this experiment, we compare the FQRC with the state-of-the-art multi-label scene classification approaches
\cite{Boutell_Luo_Shen_Brown_2004,Zhang_2007}. This comparison is
performed with MLS dataset. The comparison is done on two aspects:
computational complexity and accuracy.

\subsubsection{\textbf{Computational Complexity}}

First, we show the complexity of our method compare to both approaches
with the results presented in Table \ref{compx}. In this context, $N$
denotes the number of classes, $M$ is the number of features, and $T$
is the number of data. The training complexity of \cite{Zhang_2007}
consists of three parts; prior, conditional probability, and the main
function of training, while \cite{Boutell_Luo_Shen_Brown_2004}
requires to train a classifier for every base class. These greatly
increase the computational cost compare to the FQRC.

\begin{table}[ht]
	\centering
		\caption{Complexity of FQRC compared to \cite{Boutell_Luo_Shen_Brown_2004} and \cite{Zhang_2007}}
		\label{compx}
{\renewcommand{\arraystretch}{1.5}
		\begin{tabular}{|c|c|c|}
					\hline
		  		\multirow{2}{*}{Method} & \multicolumn{2}{|c|}{Part}\\ \cline{2-3} & Training Phase & Testing Phase \\ \hline
		  		\cite{Zhang_2007} & $O(N)+O(T)+(O(3TN)+O(N))$ & $O(2N)$ \\ \hline
				\cite{Boutell_Luo_Shen_Brown_2004} & $O(NT^3)$ & $O(N)$ \\ \hline		
				\textbf{FQRC} & $O(NM)$ & $O(NM)$	\\ \hline
				
		\end{tabular}}
\end{table}

In order to verify the complexity of these methods, the computational
time comparison is done with the results show in Table
\ref{comptime}. From the result, we notice that, our method use the
shortest time to train the model which is almost 6x faster than
\cite{Zhang_2007} and 227x faster than
\cite{Boutell_Luo_Shen_Brown_2004}. However, our inference takes a
longer time compared to both methods. This is because we retrieve the
fuzzy membership values by considering all the classes of 4-tuples
membership functions that corresponds to all the features. This also
means that with a reduction in terms of the number of features, we can
obtain faster computational speed. The computational time result for
testing is done using all the testing data, so it is still acceptable if we apply only one data per cycle with an average of 3 milliseconds of computational time. Nonetheless, \cite{Zhang_2007} suffered from finding the optimal number of nearest neighbor involved in the classification step. This had directly affects the performance of the classification. 

\begin{table}[htbp]
	\centering
		\caption{Computational time of FQRC compared to \cite{Boutell_Luo_Shen_Brown_2004} and \cite{Zhang_2007} on MLS dataset}
		\label{comptime}
{\renewcommand{\arraystretch}{1.5}
		\begin{tabular}{|c|c|c|c|}
			\hline
  		\multirow{2}{*}{Method} & \multicolumn{3}{|c|}{Computational time (s)}\\ \cline{2-4}
  		 & Training & Testing & Overall  \\ \hline
  		\cite{Zhang_2007} & 0.9363 & 0.5662 & \textbf{1.5025} \\ \hline
		\cite{Boutell_Luo_Shen_Brown_2004} & 37.8859 & \textbf{0.3725} & 38.2584	\\ \hline		
		\textbf{FQRC} & \textbf{0.1666} & 3.9479	& 4.1145 \\ \hline
		\end{tabular}}
\end{table}

\subsubsection{\textbf{Accuracy}}

For fair comparison, instead of employing all the scene data from the
MLS scene dataset, we only selected the multi-label class scene data. It means we eliminate those testing data that are
categorized as base class in \cite{Boutell_Luo_Shen_Brown_2004}
according to the ground truth and use only the test data in
multi-label class. This explains why the results are different from
the original paper. Again, we should point out that the intention of
this work is focused on the multi-label scene classification.

\begin{table}[htbp]
	\centering
		\caption{$\alpha$-Evaluation of FQRC compared to \cite{Zhang_2007} and \cite{Boutell_Luo_Shen_Brown_2004}}
		\label{table:eva}
{\renewcommand{\arraystretch}{1.1}
\resizebox{7cm}{!} {
		\begin{tabular}{|c|c|c|c|c|}
			\hline
  		\multirow{2}{*}{Method} & \multicolumn{4}{|c|}{$\alpha$-evaluation}\\ \cline{2-5}
  		 & $\alpha=0$ & $\alpha=0.5$ & $\alpha=1$ & $\alpha=2$  \\ \hline
  		 \cite{Zhang_2007} & 1 & 0.54 & 0.39 & 0.20 \\ \hline
		\cite{Boutell_Luo_Shen_Brown_2004} & 1 & 0.69 & 0.49 & 0.27	\\ \hline		
		\textbf{FQRC} & \textbf{1} & \textbf{0.69}	& \textbf{0.54} & \textbf{0.37} \\ \hline
		
		\end{tabular}}
		}
\end{table}

Based on \cite{Boutell_Luo_Shen_Brown_2004}, $\alpha$ is the
forgiveness rate because it reflects how much to forgive the errors made
in predicting labels. Small value of $\alpha$ is more aggressive (tend
to forgive error) while a high value is conservative (penalizing error
more harshly). In relation to the multi-label classification,  $\alpha
= \infty$ with a score = 1 occurs only when the prediction is fully
correct (all hit and no missed) or 0 otherwise. On the other hand,
when $\alpha = 0$, we get the score = 1 except when the
answer is fully incorrect (all missed). From Table \ref{table:eva}, we could
observe that the FQRC outperforms the two other methods with better
accuracy in the $\alpha$-evaluation.

In summary, we have tested the performances of FQRC compared to
\cite{Boutell_Luo_Shen_Brown_2004,Zhang_2007} using MLS scene dataset
and obtained superior results. The key factors which distinguish our
work from them include: Firstly, we do not require the human
intervention in manually annotate the multi-label class images to
serve as prior information. This is impractical because it may lead to
a large number of classes with sparse sample in the dataset
\cite{tsoumakas2007multi}. For instance, the class name ``Field + Fall
foliage + Mountain'' has only one image in
\cite{Boutell_Luo_Shen_Brown_2004}. Secondly, human annotation is
bias, that is different people from different background tend to
provide different answers for a scene image. We showed this scenario
in our real-world online survey results as well as psychological and
metaphysical studies \cite{Forguson}. Thirdly,
\cite{Boutell_Luo_Shen_Brown_2004,Zhang_2007} only output binary results in multi-label classification task while our proposed approach provides ranking information.

\section{Conclusion}
\label{Con}
In this paper, we raised an important issue that scene images are
non-mutually exclusive. Unfortunately, almost all the existing works
focused on scene image understanding assumed that images are mutually
exclusive. Related works that do not perform this such as in
\cite{Boutell_Luo_Shen_Brown_2004, Zhang_2007} employed human expert
to re-label the image manually in order to obtain multi-label class
training data for further processing, which is impractical and
bias. Our aim is to raise the awareness in the community regarding
this very important, but largely neglected issue. In order to achieve
this, we conducted an online survey among people from the different
background and subsequently proposed the ranking classifier (FQRC)
which adopting fuzzy qualitative principle as a resolution. The
results from extensive experiments have shown the effectiveness,
feasibility, and efficiency of our proposed approach as compared to
the other state-of-the-art approaches. Our future work will focus on
extending the work with the use a fuzzy loss function \cite{Pagola} and normalized
sum of memberships, as well as the investigate the effects of different membership
function as the learning model.


\bibliographystyle{IEEEtran}
\bibliography{TFS2012}

\end{document}